\documentclass[lettersize,journal]{IEEEtran}
\usepackage{algorithmic}
\usepackage{algorithm}
\usepackage{array}
\usepackage[caption=false,font=scriptsize,labelfont=sf,textfont=sf]{subfig}
\usepackage{textcomp}
\usepackage{stfloats}
\usepackage{url}
\usepackage{amsmath,amssymb,amsfonts,mathtools,bm}
\usepackage{verbatim}
\usepackage{graphicx}
\usepackage{cite}
\usepackage{hyperref}
\usepackage{booktabs}
\usepackage{tikz}
\usetikzlibrary{arrows.meta,calc,positioning,bending}

\begin{document}

\title{Out-of-Distribution Radar Detection with Complex VAEs:
Theory, Whitening, and ANMF Fusion}

\author{Y.~A.~Rouzoumka,
        J.~Pinsolle,
        E.~Terreaux,
        C.~Morisseau,
        J.-P.~Ovarlez,
        and C.~Ren%
\thanks{Y.~A.~Rouzoumka, E.~Terreaux, C.~Morisseau, and J.-P.~Ovarlez are with DEMR, ONERA, Universit\'e Paris-Saclay, F-91123 Palaiseau, France (e-mail: \texttt{firstname.lastname@onera.fr}), .}
\thanks{Y.~A.~Rouzoumka, J.~Pinsolle, J.-P.~Ovarlez, and C.~Ren are with SONDRA, CentraleSup\'elec, Universit\'e Paris-Saclay, 91190 Gif-sur-Yvette, France (e-mail: \texttt{firstname.lastname@centralesupelec.fr}).}%
}

\markboth{IEEE Transactions on Signal Processing}%
{Rouzoumka \MakeLowercase{\textit{et al.}}: Out-of-Distribution Radar Detection with Complex VAEs}


\maketitle

\begin{abstract}
We investigate the detection of weak complex-valued signals immersed in non-Gaussian, range-varying interference, with emphasis on maritime radar scenarios. The proposed methodology exploits a Complex-valued Variational AutoEncoder (CVAE) trained exclusively on clutter-plus-noise to perform Out-Of-Distribution detection. By operating directly on in-phase/quadrature samples, the CVAE preserves phase and Doppler structure and is assessed in two configurations: (i) using unprocessed range profiles and (ii) after local whitening, where per-range covariance estimates are obtained from neighboring profiles. Using extensive simulations together with real sea-clutter data from the CSIR maritime dataset, we benchmark performance against classical and adaptive detectors (MF, NMF, AMF-SCM, ANMF-SCM, ANMF-Tyler). In both configurations, the CVAE yields a higher detection probability $P_d$ at matched false-alarm rate $P_{fa}$, with the most notable improvements observed under whitening. We further integrate the CVAE with the ANMF through a weighted $\log-p$ fusion rule at the decision level, attaining enhanced robustness in strongly non-Gaussian clutter and enabling empirically calibrated $P_{fa}$ control under $H_0$. Overall, the results demonstrate that statistical normalization combined with complex-valued generative modeling substantively improves detection in realistic sea-clutter conditions, and that the fused CVAE–ANMF scheme constitutes a competitive alternative to established model-based detectors.
\end{abstract}


\section{Introduction}
\label{sec:intro}
\IEEEPARstart{D}{etection} of weak complex-valued signals immersed in non-Gaussian interference is a central problem in statistical signal processing~\cite{Greco16}, with applications ranging from radar and sonar to wireless communications and passive sensing. In many of these modalities, the background exhibits range- or time-varying statistics, heavy tails, and residual impropriety after coherent processing, so that classical Gaussian-based detectors struggle to maintain high detection probability and reliable false-alarm rate control.

In radar target detection, the goal is to discriminate targets from complex clutter environments such as maritime or airborne scenes. Classical detectors, such as the Matched Filter (MF), the Normalized Matched Filter (NMF), and adaptive approaches including the Adaptive Matched Filter (AMF)~\cite{Robey1992ACA}, Kelly’s GLRT~\cite{4104190}, and the Adaptive Normalized Matched Filter (ANMF)~\cite{301849}, achieve strong performance under homogeneous complex Gaussian clutter or well-specified compound-Gaussian models. Their effectiveness deteriorates, however, when clutter exhibits pronounced non-Gaussian characteristics combined with thermal noise, making it difficult to preserve a good $P_d$-$P_{\mathrm{fa}}$ trade-off.

Recent advances in deep learning have opened new perspectives for radar detection~\cite{GoodBengCour16}. Variational Autoencoders (VAEs) are particularly appealing for out-of-distribution (OOD) detection~\cite{ran2021,yang2024} thanks to their ability to model complex data distributions~\cite{Kingma_2019} and to flag deviations from learned background statistics. Beyond radar, VAEs and related deep generative models have been used for acoustic monitoring, medical imaging, and high-voltage equipment diagnostics~\cite{MRAMJ2022,BS2023,marimont2020,MSWPNM2021}. In radar applications, a few recent works address specific short-range FMCW scenarios~\cite{KSS2023}, but the systematic integration of complex VAEs into classical CFAR detection pipelines is largely unexplored.

Building on these advances, this paper introduces a complex-valued VAE (CVAE) architecture that operates directly on complex radar signals~\cite{rouzoumkacvae2025}, thereby preserving phase information and increasing flexibility in the latent space. Complex-valued neural networks exploit joint amplitude–phase representations and have shown advantages in several signal processing tasks~\cite{fixtorchcvnn,barrachina:hal-03771786,complexreccurrxie,deepcomplexconvxiangyang}. Here, the CVAE is used as a generative model of clutter profiles; detection relies on a complex reconstruction-error statistic, which is calibrated via per-Doppler probability-integral transforms (PIT) to enforce CFAR. We also consider a local covariance-based whitening that normalizes each Doppler profile using a neighborhood-based covariance estimate, aiming to stabilize the null distribution of CVAE scores.

To connect learned and model-based detectors, we develop a decision-level fusion strategy combining the CVAE with the classical ANMF. Both detectors are mapped to a common significance scale expressed as $p$-values under $H_0$, and then fused via a weighted logarithmic combination reminiscent of Fisher-type $p$-value aggregation~\cite{Fisher1932,Lancaster1961}. This enables the ANMF to dominate in well-modelled Gaussian or mildly heterogeneous regimes, while the CVAE contributes robustness when clutter exhibits strong non-Gaussian behavior, with the fusion designed to preserve CFAR via empirical calibration.

In addition to extensive simulation studies with correlated Gaussian and compound-Gaussian clutter, the proposed framework is validated on real maritime radar data from the CSIR sea-clutter dataset~\cite{deWind2010DataWare,Herselman2007Analysis}. These measurements contain sea-surface returns and small targets under varying sea states and operating conditions, and have been used to study wave-height estimation and deep-learning-based sea-clutter modelling~\cite{Liu2021WaveHeight,Ma2019SeaClutterDL,Gong2020LightCNN,Ding2020MarineDataset,Tan2025GNNSeaClutter,Kandagatla2025NNSeaClutter}, providing a realistic testbed to assess the robustness of the CVAE-based detector and its fusion with ANMF.

\textbf{Contributions.}
This paper makes three main contributions.

(i) We formulate weak-signal detection in heterogeneous, non-Gaussian complex backgrounds as a complex-valued OOD detection problem, and we propose a CVAE architecture that operates directly on complex Doppler profiles. The latent posterior is modeled as a diagonal non-circular complex Gaussian, and we derive a closed-form KL divergence to a circular complex prior, generalizing standard real-valued and circular complex VAE formulations.

(ii) We study the impact of local covariance-based whitening as a generic statistical normalization for non-stationary complex-valued data. By comparing raw and locally whitened regimes on correlated Gaussian and compound-Gaussian clutter, as well as on real sea clutter, we show that combining local whitening with complex generative modeling can significantly sharpen detection fronts at fixed false-alarm rate compared with classical CFAR detectors.

(iii) We introduce a decision-level fusion scheme that combines a data-driven CVAE detector with a model-based ANMF through weighted log-$p$ aggregation. Both branches are calibrated via PIT under $H_0$, and the fused statistic is thresholded with an empirical CFAR-like rule. This $p$-value fusion framework remains valid under dependence and yields detectors that inherit robustness in heavy-tailed clutter while preserving controllable false-alarm behaviour. Although our experiments focus on maritime radar, the overall framework applies to a broad class of complex-valued sensing systems, including sonar, passive radar, and MIMO communications.

The remainder of this paper is organized as follows. Section~\ref{sec:statmodel} reviews statistical models of radar clutter and classical detection benchmarks. Section~\ref{sec:ood_overview} discusses deep learning-based OOD detection and its specific challenges in radar. Section~\ref{sec:cvae_detector} presents the proposed CVAE-based detector, including the complex-valued architecture, local whitening, and CVAE–ANMF fusion. Section~\ref{sec:results} details the training methodology and experimental results. Section~\ref{sec:conclu} concludes the paper.

\textit{Notation:} Matrices are denoted by bold uppercase letters and vectors by bold lowercase letters. For any matrix or vector $\mathbf{A}$, $\mathbf{A}^T$ and $\mathbf{A}^H$ denote the transpose and Hermitian (conjugate) transpose, respectively. $\mathbf{I}$ represents the identity matrix. We denote respectively by $\mathcal{N}(\boldsymbol{\mu}, \boldsymbol{\Gamma})$ and $\mathcal{CN}(\boldsymbol{\mu}, \boldsymbol{\Gamma})$ real-valued and complex circular Gaussian distributions with mean $\boldsymbol{\mu}$ and covariance $\boldsymbol{\Gamma}$, respectively. And $\mathcal{N}(\boldsymbol{\mu}, \boldsymbol{\Gamma}, \boldsymbol{\Delta} )$ denotes complex (non circular) Gaussian distribution with mean $\boldsymbol{\mu}$ and covariance $\boldsymbol{\Gamma}$ and pseudo-covariance $\boldsymbol{\Delta}$ 
The operator $\boldsymbol{\mathcal{T}}(\rho)$ denotes a Toeplitz matrix constructed from a correlation coefficient $\rho$, with $\{\boldsymbol{\mathcal{T}}(\rho)\}_{i,j} = \rho^{|i-j|}$. The symbols $\odot$ and $\oslash$ denote the Hadamard (element-wise) product and division, respectively. Similarly, $|\cdot|^\circ$, $\cdot^{\circ \alpha}$ and $\log^\circ(\cdot)$ indicate the element-wise absolute value, power to the exponent $\alpha$, and logarithm of a vector. The operators $\Re\{\cdot\}$ and $\Im\{\cdot\}$ extract the real and imaginary parts of a complex argument. The operator $\mathrm{diag}(\mathbf{.})$ extracts the diagonal elements of a matrix.

\section{Statistical Model}
\label{sec:statmodel}

\subsection{Signal Model and Binary Hypotheses}
\label{ssec:hypothesis}
In adaptive radar detection, we test for the presence of a complex target return $\alpha\,\mathbf{p}\in\mathbb{C}^m$ (with unknown complex amplitude $\alpha$ and known steering vector $\mathbf{p}$ of dimension $m$) embedded in background clutter $\mathbf{c}$ and independent white thermal noise $\mathbf{n}\sim\mathcal{CN}(\mathbf{0},\sigma^2\mathbf{I})$. The problem is cast as
\begin{equation}
\begin{cases}
H_0:\; \mathbf{z}=\mathbf{c}+\mathbf{n}, & \text{(no target)},\\[2pt]
H_1:\; \mathbf{z}=\alpha\,\mathbf{p}+\mathbf{c}+\mathbf{n}, & \text{(target present)},
\end{cases}
\end{equation}
where $\mathbf{z}\in\mathbb{C}^m$ is the received snapshot. In homogeneous environments, the clutter is modeled as circular complex Gaussian $\mathbf{c}\sim\mathcal{CN}(\mathbf{0},\boldsymbol{\Sigma}_c)$. In heterogeneous environments a compound-Gaussian model is used, $\mathbf{c}=\sqrt{\tau}\,\mathbf{g}$ with $\mathbf{g}\sim\mathcal{CN}(\mathbf{0},\boldsymbol{\Sigma}_c)$ and a positive texture parameter $\tau>0$ describing pulse-to-pulse power fluctuations; conditionally on $\tau$, $\mathbf{c}\sim\mathcal{CN}(\mathbf{0},\tau\,\boldsymbol{\Sigma}_c)$, and we fix the power scale via $\mathbb{E}[\tau]=1$. After whitening by the total covariance $\boldsymbol{\Sigma}=\boldsymbol{\Sigma}_c+\sigma^2\mathbf{I}$, the signal-to-noise ratio under $H_1$ reads
\begin{equation}
\mathrm{SNR} = |\alpha|^2\,\mathbf{p}^H\boldsymbol{\Sigma}^{-1}\mathbf{p}.
\end{equation}


\subsection{Classical Detectors}
When $\boldsymbol{\Sigma}$ is known, the MF is GLRT-optimal in Gaussian environments, and the associated detection test is expressed as:
\begin{equation}
\Lambda_{\text{MF}}(\mathbf{z})=\frac{\bigl|\mathbf{p}^H\boldsymbol{\Sigma}^{-1}\mathbf{z}\bigr|^2}{\mathbf{p}^H\boldsymbol{\Sigma}^{-1}\mathbf{p}}
\;\underset{H_0}{\overset{H_1}{\gtrless}}\;\lambda,
\label{eq:MF}
\end{equation}
with $\Lambda_{\text{MF}}$ exponentially distributed with unit mean under $H_0$ (equivalently $\Lambda_{\text{MF}}\overset{d}{=}\displaystyle\frac{1}{2}\chi^2_2$ where $\chi^2_2$ stands for Chi-Square Distribution with two degrees of freedom), yielding $P_{fa}=e^{-\lambda}$ in the homogeneous Gaussian case. In partially homogeneous noise, where the covariance is known up to an unknown positive scale, the scale-invariant NMF \cite{301849} is defined as:
\begin{equation}
\Lambda_{\text{NMF}}(\mathbf{z})=
\frac{\bigl|\mathbf{p}^H\boldsymbol{\Sigma}^{-1}\mathbf{z}\bigr|^2}{\bigl(\mathbf{p}^H\boldsymbol{\Sigma}^{-1}\mathbf{p}\bigr)\bigl(\mathbf{z}^H\boldsymbol{\Sigma}^{-1}\mathbf{z}\bigr)}
\;\underset{H_0}{\overset{H_1}{\gtrless}}\;\lambda.
\label{eq:NMF}
\end{equation}
These MF/NMF benchmarks are well understood and effective when clutter plus thermal noise remain Gaussian.

\subsection{Adaptive Detection}
In practice,  $\boldsymbol{\Sigma}$ is unknown and has to be estimated from $K$ secondary data $\{\mathbf{z}_k\}_{k=1}^K$ supposed to be target-free. In Gaussian environments, the best estimator of $\boldsymbol{\Sigma}$ is the Sample Covariance Matrix (SCM)
\begin{equation}
\widehat{\boldsymbol{\Sigma}}_{\text{SCM}}=\frac{1}{K}\sum_{k=1}^K \mathbf{z}_k\mathbf{z}_k^H,
\end{equation}
leading to the Adaptive Matched Filter (AMF-SCM)~\cite{Robey1992ACA} and the Adaptive Normalized Matched Filter (ANMF-SCM)~\cite{Kraut2001AdaptiveSD}, obtained by inserting $\widehat{\boldsymbol{\Sigma}}_{\text{SCM}}$ into~\eqref{eq:MF} and~\eqref{eq:NMF}. In compound-Gaussian (impulsive) clutter, SCM-based detectors may lose CFAR properties and degrade in $P_d$. Robust covariance estimation is then preferable such as $M$-estimators~\cite{tyler1987,pascal08,6263313,7383755,Pascal8} and Tyler’s fixed-point (FP) estimator. The latter is defined implicitly by
\begin{equation}
\widehat{\boldsymbol{\Sigma}}_{\text{FP}}=\frac{m}{K}\sum_{k=1}^K
\frac{\mathbf{z}_k\mathbf{z}_k^H}{\mathbf{z}_k^H \widehat{\boldsymbol{\Sigma}}_{\text{FP}}^{-1}\mathbf{z}_k},
\end{equation}
which is scale invariant at the snapshot level (only the normalized snapshots matter) and thus attenuates large texture fluctuations. The resulting ANMF-Tyler is obtained by replacing $\boldsymbol{\Sigma}^{-1}$ with $\widehat{\boldsymbol{\Sigma}}_{\text{FP}}^{-1}$ in~\eqref{eq:NMF}; it is known to be effective and more stable in heavy-tailed clutter~\cite{tyler1987,pascal08,6263313,7383755,Pascal8}. However, once additive thermal noise is present, the returns no longer follow a pure compound-Gaussian (texture$\times$speckle) model. Moreover, this breaks the texture invariance property of Tyler’s estimator and can degrade CFAR control, even when a robust covariance estimator is used. 

\subsection{Real-World Challenges and Motivation for Learning-Based OOD}
Sea clutter is markedly non-Gaussian with pronounced Range slow-time correlations; combining such clutter with thermal noise yields settings where no closed-form optimal detector is known. Classical Gaussian-based detectors (MF, AMF-SCM, ANMF-SCM) remain appropriate when the overall disturbance is Gaussian, but performance and false-alarm regulation can be severely affected otherwise. Robust extensions that plug Tyler’s M-estimator in place of the SCM (ANMF-Tyler) restore CFAR behaviour under compound-Gaussian clutter, but they still rely on locally homogeneous training data and can degrade in heterogeneous clutter plus white-noise regimes. These limitations motivate data-driven alternatives. In the remainder, we leverage an out-of-distribution approach based on a CVAE trained on target-free data, and assess its use both on raw profiles and after local whitening; we further consider decision-level fusion with a model-based detector to capitalize on their complementary strengths.

\section{Deep Learning-Based OOD Detection for Radar}
\label{sec:ood_overview}

\subsection{The Need for OOD Detection in Radar Processing}
\label{ssec:ood_importance}
Classical radar detectors, such as MF, NMF, AMF, and ANMF, are derived within the Neyman-Pearson likelihood ratio framework \cite{neyman1933problem} under explicit statistical models for the disturbance (clutter plus thermal noise), typically Gaussian or compound-Gaussian~\cite{Robey1992ACA,Kraut2001AdaptiveSD}. Their optimality hinges on how well these statistical assumptions match reality. In operational settings (maritime surveillance, airborne GMTI, coastal monitoring), clutter is heterogeneous, non-stationary, and structurally rich; it exhibits long-range correlations, intermittency, and heavy tails that vary across range–Doppler cells and from scan to scan. Under such departures, even well-engineered two-step adaptive schemes suffer: the number of target-free secondary data may be insufficient or contaminated by outliers, covariance estimates become unreliable, and the false-alarm rate regulation deteriorates.

OOD detection addresses a different question than classical GLRTs: rather than testing a parametric signal-in-noise model, it asks whether a given return is statistically compatible with the background distribution learned from target-free data. This data-driven view avoids committing to a potentially misspecified analytic clutter model and instead lets the detector adapt directly to the empirical site and sea-state–dependent statistics~\cite{yang2024}. In other words, OOD detection provides a principled way to flag anomalies (putative targets) when explicit modeling is intractable or unreliable.

\subsection{Deep Learning Approaches to OOD}
\label{ssec:ood_methods}
Modern deep OOD methods fall into three broad (and complementary) families.  
\emph{Reconstruction-based} approaches train autoencoders (AEs) or VAEs to reconstruct in-distribution (ID) data with small error~\cite{papierICASSP}, with the idea that OOD samples will incur a larger distortion because they are not well represented by the learned manifold~\cite{Baur_2021,ran2021}.
For a sample $\mathbf{z}$ and reconstruction $\hat{\mathbf{z}}$, the reconstruction squared error  $\mathcal{L}_{\text{rec}}=\|\mathbf{z}-\hat{\mathbf{z}}\|^2$ serves as an anomaly score, and thresholds are chosen to meet a prescribed $P_{fa}$. 
\emph{Density-estimation} methods seek calibrated likelihoods under expressive generative models. 
Normalizing flows and related hybrids (e.g., SurVAE) learn invertible mappings that transform complex data to a tractable base density, so that likelihoods can be evaluated exactly via the change-of-variables formula~\cite{NEURIPS2020_41d80bfc,NEURIPS2020_9578a63f}. Energy-based objectives offer another route to likelihood-based OOD scoring~\cite{NEURIPS2020_f5496252}.  
\emph{Representation-learning/contrastive} methods shape a feature space where ID clusters tightly and anomalies fall away; decisions can be formed using distances such as the Mahalanobis metric computed from ID statistics~\cite{zhou-etal-2021-contrastive,9761434}. 

In radar, reconstruction-based methods are attractive because they are unsupervised, computationally efficient at test time, and naturally compatible with stringent $P_{fa}$ control, while avoiding the calibration challenges that plague likelihoods in very high dimensions.

\subsection{OOD in Radar: State of the Art and Gaps}
\label{ssec:ood_radar}
Evidence for the effectiveness of OOD has accumulated across various domains. In medical imaging, likelihood modeling in compressed latent spaces has enabled state-of-the-art anomaly localization~\cite{pmlr-v172-graham22a}. For time series, attention-based architectures (Anomaly Transformer, TranAD) capture long-range dependencies and highlight subtle deviations~\cite{Xu2021AnomalyTT,tuli2022tranad}. In radar specifically, reconstruction-based OOD detectors on short-range
60\,GHz FMCW sensors have shown that autoencoder reconstruction errors
(and related latent-energy scores) correlate strongly with anomalous
micro-Doppler patterns~\cite{HOOD}, and our preliminary results indicate similar benefits for maritime returns~\cite{rouzoumkacvae2025}. However, most prior studies operate in controlled or short-range settings, leaving open the question of handling strongly correlated, heavy-tailed sea clutter with additive thermal noise, precisely where classical detectors struggle.

\subsection{Reconstruction-Driven CVAE for Radar OOD}
\label{ssec:vae_radar}
We adopt a complex-valued variational autoencoder ~\cite{rouzoumkacvae2025} as the backbone of our OOD model. Training is performed exclusively on target-free radar profiles (clutter plus thermal noise), allowing the encoder-decoder pair to learn a compact latent representation of the background. Test samples that contain a target then deviate from this manifold by inducing a larger reconstruction error $\mathcal{L}_{\text{rec}}=\|\mathbf{z}-\hat{\mathbf{z}}\|^2 =\sum_{n}|z_n-\hat{z}_n|^2 $, which we use as our primary anomaly score in this work. In principle, one could also exploit the CVAE latent posterior as an additional signal, for instance by penalizing low-probability regions of the variational family. 

Crucially, the network is complex-valued end-to-end, ingesting complex-valued data and maintaining phase coherence through complex convolutions, normalizations, and activations. This is not merely an architectural convenience: phase carries essential information for coherent integration and Doppler discrimination. Empirically and theoretically, complex-valued networks can dominate real-valued counterparts at fixed capacity in phase-sensitive modalities such as Polarimetric SAR classification~\cite{barrachina:hal-03771786}, and recent generative designs show that complex VAEs capture structure in noisy, phase-rich signals more faithfully than real-valued surrogates~\cite{rouzoumkacvae2025,complexreccurrxie,deepcomplexconvxiangyang}. We therefore expect CVAEs to better preserve the geometry of sea clutter and the subtle distortions induced by weak targets.

\section{Complex VAE Radar Detector: Architecture, Whitening, and Fusion}
\label{sec:cvae_detector}

Parametric statistical modeling of sea clutter plus thermal noise is fragile under heterogeneity and non-Gaussianity. We therefore learn the null ($H_0$) distribution from a target-free corpus $\mathcal{D}_{H_0}=\{\mathbf{z}_n\}_{n\in[1,N]}$ using a Complex-valued VAE. This choice offers a pragmatic compromise for radar: the CVAE is fully unsupervised, operates natively in $\mathbb{C}$ to preserve Doppler-phase cues, and yields a reconstruction-based anomaly score that can be calibrated to a target false-alarm rate $P_{fa}$.
Beyond raw CVAE processing, we consider local whitening (Section~\ref{ssec:datawhitening}) to stabilise clutter statistics, and a decision-level fusion with ANMF via weighted log-$p$ combination (Section~\ref{ssec:fusion}) to leverage complementary inductive biases across the Gaussian-to-impulsive spectrum.

\subsection{Complex-Valued VAE Architecture}
\label{ssec:ourvae}

\begin{figure*}[!t]
\centering
\includegraphics[width=1.5\columnwidth]{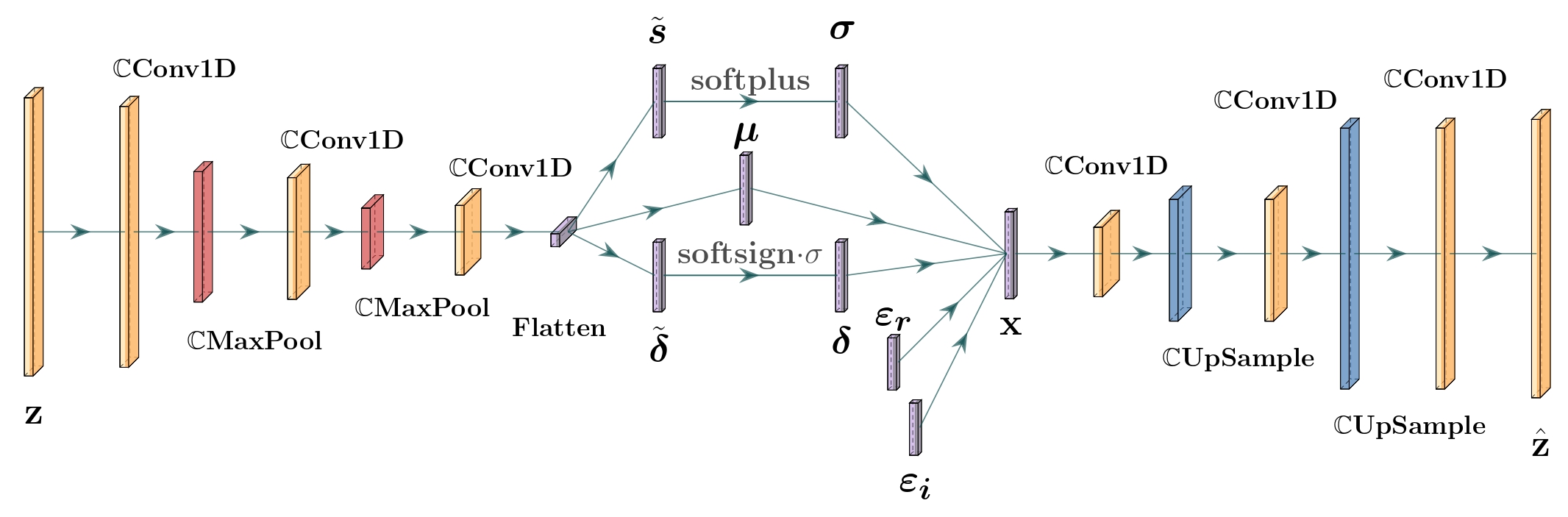}%
\caption{Complex-valued VAE architecture used in this work. The encoder applies complex 1D convolutions and pooling to a Doppler profile $\mathbf{z}\in\mathbb{C}^m$ and produces three heads $(\boldsymbol{\mu},\tilde{\boldsymbol{s}},\tilde{\boldsymbol{\delta}})$. The unconstrained outputs $\tilde{\boldsymbol{s}}$ and $\tilde{\boldsymbol{\delta}}$ are mapped to the variance $\boldsymbol{\sigma}$ and pseudo-variance $\boldsymbol{\delta}$ via a softplus and a complex softsign$\cdot\boldsymbol{\sigma}$ transform. A complex reparameterization with two real noise vectors $\boldsymbol{\varepsilon}_r,\boldsymbol{\varepsilon}_i$ yields latent samples $\mathbf{x}\in\mathbb{C}^q$, which are decoded back to $\hat{\mathbf{z}}\in\mathbb{C}^m$ via complex transposed convolutions and upsampling.}
\label{fig:vae}
\end{figure*}

\paragraph*{Design principle}
Operating natively in $\mathbb{C}$ preserves phase coherence and enables modeling of non-circular effects that arise after coherent processing (e.g., sea clutter with residual impropriety). Our CVAE therefore operates directly on complex-valued Doppler profiles $\mathbf{z}\in\mathbb{C}^{m}$, using complex convolutions, complex pooling, and complex activation functions throughout the encoder and decoder (Fig.~\ref{fig:vae}). This avoids the information loss associated with ad-hoc real-imaginary stacking, and follows recent evidence that complex-valued networks can outperform real-valued ones when phase is discriminative~\cite{barrachina:hal-03771786,complexreccurrxie,deepcomplexconvxiangyang}.

\paragraph*{Encoder}
Given a Doppler profile $\mathbf{z}\in\mathbb{C}^{m}$ (raw or whitened, cf.\ Sec.~\ref{ssec:datawhitening}), a stack of $K$ complex 1D convolutional blocks maps
\[
\mathbf{z} \;\longmapsto\; \mathbf{h}= \mathrm{Conv1Dblock(\mathbf{z})}\in\mathbb{C}^{C_K\times m_K},
\]
where each block uses a complex convolution, a complex activation (modReLU or CReLU), optional complex max-pooling (implemented as max-pooling on $|\cdot|$ with complex gather), and complex batch-normalization.
The terminal feature tensor $\mathbf{h}$ is flattened and projected to three unconstrained heads
\[
\boldsymbol{\mu}\in\mathbb{C}^q,\qquad
\tilde{\boldsymbol{s}}\in\mathbb{R}^q,\qquad
\tilde{\boldsymbol{\delta}}\in\mathbb{C}^q.
\]
We write $\boldsymbol{\sigma}=\mathrm{softplus}(\tilde{\boldsymbol{s}})\in\mathbb{R}_+^q$ for the element-wise standard deviation, so that the variance in the latent distribution is $\boldsymbol{\sigma}^{\circ 2}$, and we obtain the pseudo-variance $\boldsymbol{\delta}$ from $\tilde{\boldsymbol{\delta}}$ via the constraint below.
Taken together, $(\boldsymbol{\mu},\boldsymbol{\sigma}^{\circ 2},\boldsymbol{\delta})$ parameterize a non-circular complex Gaussian latent of dimension $q$; this family strictly generalizes the circular case ($\boldsymbol{\delta}=\mathbf{0}$).

\paragraph*{Stable constraint for the pseudo-variance}
Let $(\tilde{\boldsymbol{s}},\tilde{\boldsymbol{\delta}},\boldsymbol{\mu})$ denote the unconstrained encoder outputs.
We enforce numerically stable element-wise constraints via
\begin{equation}
\label{eq:cvae-sigma-delta}
\boldsymbol{\sigma}=\mathrm{softplus}(\tilde{\boldsymbol{s}}),\qquad
\boldsymbol{\delta}=\boldsymbol{\eta}\odot \boldsymbol{\sigma},
\qquad
\boldsymbol{\eta}=\frac{\tilde{\boldsymbol{\delta}}}{1+|\tilde{\boldsymbol{\delta}}|^\circ},
\end{equation}
where $|\cdot|^\circ$ denotes the element-wise complex modulus and $\odot$ is the Hadamard product.
The "complex softsign" mapping
\[
|\boldsymbol{\eta}|^\circ
=\frac{|\tilde{\boldsymbol{\delta}}|^\circ}{1+|\tilde{\boldsymbol{\delta}}|^\circ}
<1
\]
guarantees $|\boldsymbol{\delta}|^\circ<\boldsymbol{\sigma}$ by construction, while preserving the phase of $\tilde{\boldsymbol{\delta}}$.
In implementation, we add a small $\varepsilon>0$ to $\boldsymbol{\sigma}$ and to the denominators in the reparameterization for numerical robustness, but we omit $\varepsilon$ from the formulas for clarity.

\paragraph*{Complex reparameterization}
Given $(\boldsymbol{\mu},\boldsymbol{\sigma},\boldsymbol{\delta})$, we sample a latent code $\mathbf{x}\in\mathbb{C}^q$ via
\begin{equation}
\label{eq:cvae-reparam}
\mathbf{x}
= \boldsymbol{\mu}
+ \boldsymbol{k}_r \odot \boldsymbol{\epsilon}_r
+ i\, \boldsymbol{k}_i \odot \boldsymbol{\epsilon}_i\, ,
\qquad
\boldsymbol{\epsilon}_r,\boldsymbol{\epsilon}_i \overset{\text{i.i.d.}}{\sim} \mathcal{N}(\mathbf{0},\mathbf{I})\, ,
\end{equation}
where the element-wise scaling vectors $\boldsymbol{k}_r,\boldsymbol{k}_i\in\mathbb{R}_+^q$ are chosen so that
\begin{align*}
    \boldsymbol{\sigma}^{\circ 2} &= \mathrm{diag}\left( \mathbb{E}\left[(\mathbf{x} - \boldsymbol{\mu})(\mathbf{x} - \boldsymbol{\mu})^H\right]\right),
\\
 \boldsymbol{\delta} &=  \mathrm{diag}\left( \mathbb{E}\left[(\mathbf{x} - \boldsymbol{\mu})(\mathbf{x} - \boldsymbol{\mu})^\top\right]\right).
\end{align*}

Matching second-order moments leads to the element-wise mapping (cf.\ \cite{nakashika20_interspeech})
\begin{equation}
\label{eq:krki-final}
\left\{
\begin{aligned}
\boldsymbol{k}_r
&=
\frac{1}{\sqrt{2}}\,
\frac{\boldsymbol{\sigma}+\boldsymbol{\delta}}
     {\big(\boldsymbol{\sigma}+\Re(\boldsymbol{\delta})\big)^{\circ 1/2}}\,,
\\[3pt]
\boldsymbol{k}_i
&=
\frac{1}{\sqrt{2}}\,
\frac{\big(\boldsymbol{\sigma}^{\circ 2}-\big(|\boldsymbol{\delta}|^\circ\big)^{\circ 2}\big)^{\circ 1/2}}
     {\big(\boldsymbol{\sigma}+\Re(\boldsymbol{\delta})\big)^{\circ 1/2}}\, ,
\end{aligned}
\right.
\end{equation}
where all powers and absolute values are taken element-wise. In code, we add a small $\varepsilon$ inside the square roots and denominators of~\eqref{eq:krki-final} to avoid numerical issues when the denominator is small.

\paragraph*{Decoder and objective}
The decoder mirrors the encoder: a fully-connected layer reshapes $\mathbf{x}\in\mathbb{C}^q$ into feature maps, then a stack of complex 1D transposed convolutions and complex activations upsamples back to $\hat{\mathbf{z}}\in\mathbb{C}^{m}$. Reconstruction in the complex domain preserves the phase-sensitive Doppler structure exploited by the detectors.

Training minimizes over a complex-domain the well-known ELBO (Evidence Lower BOund) \cite{Kingma_2014}, which leads to the following loss function:
\begin{equation}
\label{eq:cvae-elbo}
\mathcal{L}_{\text{CVAE}}
= \underbrace{\|\mathbf{z}-\hat{\mathbf{z}}\|^2}_{\mathcal{L}_{\text{rec}}}
\;+\; \beta\,\underbrace{\mathcal{D}_{\text{KL}}\big(q(\mathbf{x}\!\mid\!\mathbf{z})\,\|\,p(\mathbf{x})\big)}_{\mathcal{L}_{\text{KL}}}\, ,
\end{equation}
where $\mathcal{L}_{\text{rec}}$ is the reconstruction term and $p(\mathbf{x})=\mathcal{CN}(\mathbf{0},\mathbf{I}_q)$ is a standard circular complex prior, and where the rate-distortion balance parameter $\beta>0$ controls the anomaly sensitivity of the detector through the Kullback-Liebler (KL) divergence between the distribution $p$ and the posterior distribution $q(\mathbf{x}\!\mid\!\mathbf{z})$ defined here as a diagonal non-circular complex Gaussian:
\[
q(\mathbf{x}\!\mid\!\mathbf{z})
=\mathcal{CN}\big(\boldsymbol{\mu},\boldsymbol{\Sigma},\boldsymbol{\Delta}\big), 
\hspace{0.1cm}
\boldsymbol{\Sigma}=\mathrm{Diag}(\boldsymbol{\sigma}^{\circ 2}), \hspace{0.1cm}
\boldsymbol{\Delta}=\mathrm{Diag}(\boldsymbol{\delta})\, ,
\]
with variance $\boldsymbol{\sigma}^{\circ 2}$ and pseudo-variance $\boldsymbol{\delta}$ parameterized as in~\eqref{eq:cvae-sigma-delta}. For this family, the KL divergence between $q(\mathbf{x}\!\mid\!\mathbf{z})$ and the circular prior admits the closed-form
\begin{equation}
\label{eq:cvae-kl-version2}
\mathcal{D}_{\text{KL}}
\;=\;
\boldsymbol{\mu}^H \boldsymbol{\mu}
+
\mathbf{1}_q^\top
\Big(
\boldsymbol{\sigma}
-
\displaystyle\frac{1}{2}\,
\log^\circ\big(
  \boldsymbol{\sigma}^{\circ 2}
  - (|\boldsymbol{\delta}|^\circ)^{\circ 2}
\big)
\Big),
\end{equation}
where $\mathbf{1}_q$ is the $q$-dimensional vector of ones. This expression reduces to the standard (complex circular) VAE KL when $\boldsymbol{\delta}=\mathbf{0}$ and, when further restricted to $\mathbb{R}$, to the usual real-valued VAE KL. A derivation based on the scalar complex Gaussian KL and the effective variance formalism~\cite{picinbono1996complexdistribution,SchreierScharf2010,nakashika20_interspeech} is provided in the online supplementary material.

\subsection{Data Preprocessing and Local Whitening}
\label{ssec:datawhitening}

We evaluate all detectors both on raw and on locally whitened data streams.
Let $\mathbf{Y}\in\mathbb{C}^{N_{\text{ranges}}\times N_{\text{pulses}}}$ denote the complex range--pulse matrix over the Doppler integration time, where $N_{\text{ranges}}$ is the number of range gates and $N_{\text{pulses}}$ the number of integrated pulses.
For each range index $r$ and short-time index $p$, we extract a slow-time snapshot
\[
  \mathbf{y}_{r,p}
  \;=\;
  \mathbf{Y}(r,p:p+m)
  \;\in\;
  \mathbb{C}^m\, ,
\]
of fixed length $m$ (matching the CVAE input dimension).

Around each Cell Under Test at range index $r$, we define a local neighborhood $\mathcal{N}_r$ of $N_{\text{adj}}$ adjacent range gates (excluding $r$ itself).
From all available slow-time snapshots in this neighborhood, we build a local data matrix
\[
  \mathbf{Y}_{\mathcal{N}_r}
  \;\in\;
  \mathbb{C}^{m\times K_r}\,,
\]
by stacking as columns the vectors $\mathbf{y}_{r',p}$ for $r'\in\mathcal{N}_r$ and all relevant snapshot indices $p$.
Here $K_r$ denotes the total number of such local snapshots.

The local sample covariance at range $r$ is then
\begin{equation}
\hat{\mathbf{R}}_{\mathcal{N}_r}
  \;=\;
  \frac{1}{K_r}\,\mathbf{Y}_{\mathcal{N}_r}\mathbf{Y}_{\mathcal{N}_r}^{H}
  \;\in\;\mathbb{C}^{m\times m}\,,
\end{equation}
which we regularize (in a Ledoit-Wolf way \cite{ledoit2004well}) by a small ridge term to avoid ill-conditioning:
\begin{equation}
\hat{\mathbf{R}}_{\text{reg}}
  \;=\;
  \hat{\mathbf{R}}_{\mathcal{N}_r}
  + \varepsilon_{\text{ridge}}\,
    \frac{\operatorname{tr}\left(\hat{\mathbf{R}}_{\mathcal{N}_r}\right)}{m}\,\mathbf{I}_m\,,
\quad
\varepsilon_{\text{ridge}}>0\,.
\end{equation}
Each slow-time snapshot at range $r$ is whitened as
\begin{equation}
\mathbf{y}_{r,p}^{\text{w}}
  \;=\;
  \hat{\mathbf{R}}_{\text{reg}}^{-\frac{1}{2}}\,\,\mathbf{y}_{r,p}\, .
\end{equation}
Finally, the corresponding one-dimensional Doppler profile is obtained by
\begin{equation}
  \mathbf{z}_{r,p}^{\text{w}}
  \;=\;
  \mathrm{DFT}\!\big(\mathbf{y}_{r,p}^{\text{w}}\big)
  \;\in\;
  \mathbb{C}^m\,.
\end{equation}
where $\mathrm{DFT}(\mathbf{v})$ stands for Discrete Fourier Transform of the vector $\mathbf{v}$.

This local CFAR-like normalization homogenizes the clutter power and reduces correlation in slow time, while the subsequent unitary DFT preserves the whitening in the Doppler domain and coherent target returns.

\begin{algorithm}[!t]
\caption{Local Whitening Before Doppler Processing}
\label{alg:whitening}
\begin{algorithmic}[1]
\REQUIRE Complex range--pulse matrix
         $\mathbf{Y}\in\mathbb{C}^{N_{\text{ranges}}\times N_{\text{pulses}}}$,
         adjacency parameter $N_{\text{adj}}$,
         ridge parameter $\varepsilon_{\text{ridge}}>0$
\ENSURE Whitened Doppler profiles $\{\mathbf{z}_{r,p}^{\mathrm{w}}\}$ (assembled into $\mathbf{Z}^{\mathrm{w}}$)
\STATE Segment $\mathbf{Y}$ into slow-time snapshots
       $\{\mathbf{y}_{r,p}\in\mathbb{C}^m\}$ for all ranges $r$ and snapshot indices $p$
       ($m$ = snapshot length).
\FOR{each range index $r = 1,\dots,N_{\text{ranges}}$}
  \STATE Define the neighborhood $\mathcal{N}_r$ of $N_{\text{adj}}$ adjacent ranges (excluding $r$).
  \STATE Form the local data matrix
         $\mathbf{Y}_{\mathcal{N}_r}
          = [\,\mathbf{y}_{r',p}\,]_{r'\in\mathcal{N}_r,\;p}
          \in\mathbb{C}^{m\times K_r}$,
         where $K_r$ is the number of local reference snapshots.
  \STATE Compute the sample covariance
         $\hat{\mathbf{R}}_r = \displaystyle\frac{1}{K_r}\mathbf{Y}_{\mathcal{N}_r}\mathbf{Y}_{\mathcal{N}_r}^{H}$.
  \STATE Regularize:
         $\hat{\mathbf{R}}_{\mathrm{reg}}
            = \hat{\mathbf{R}}_r
            + \varepsilon_{\text{ridge}}\displaystyle\frac{\operatorname{tr}(\hat{\mathbf{R}}_r)}{m}\mathbf{I}_m$.
  \FOR{each snapshot index $p$}
    \STATE Whiten slow-time snapshot:
           $\mathbf{y}_{r,p}^{\mathrm{w}} = \hat{\mathbf{R}}_{\mathrm{reg}}^{-1/2} \,\mathbf{y}_{r,p}$.
    \STATE Compute whitened Doppler profile:
           $\mathbf{z}_{r,p}^{\mathrm{w}} = \mathrm{DFT}\!\big(\mathbf{y}_{r,p}^{\mathrm{w}}\big)$.
  \ENDFOR
\ENDFOR
\STATE Assemble the whitened Doppler profiles $\{\mathbf{z}_{r,p}^{\mathrm{w}}\}$ back into a matrix $\mathbf{Z}^{\mathrm{w}}$.
\RETURN $\mathbf{Z}^{\mathrm{w}}$.
\end{algorithmic}
\end{algorithm}

\subsection{Detection Statistic and $P_{fa}$ Regulation}
\label{ssec:detectstrat}

Given a trained CVAE, the primary anomaly score is the complex reconstruction error
\begin{equation}
\label{eq:score}
s^{\text{CVAE}}(\mathbf{z})=\|\mathbf{z}-\hat{\mathbf{z}}\|^2=\sum_{n}|z_n-\hat{z}_n|^2\,.
\end{equation}
We calibrate a threshold on a \emph{disjoint} $H_0$ evaluation set to meet a target false-alarm rate:
\begin{equation}
s^{\text{CVAE}}(\mathbf{z}) \gtrless_{H_0}^{H_1} \lambda_{\text{CVAE}},\qquad 
\lambda_{\text{CVAE}} = F^{-1}_{s^{\text{CVAE}}\,|\,H_0}(1-P_{\mathrm{fa}})\, ,
\end{equation}
and we repeat the calibration for the raw and whitened streams. This empirical CFAR-like setting avoids parametric assumptions on the tail of $s^{\text{CVAE}}$ under $H_0$. From a theoretical detection perspective, the CVAE reconstruction score $s^{\text{CVAE}}(\mathbf{z})$ can be viewed as a learned, non-linear surrogate for the log-likelihood ratio between a flexible alternative model and the background model implicitly defined by the decoder. Thresholding $s^{\text{CVAE}}$ at fixed $P_{fa}$ therefore provides an empirical approximation to a Neyman-Pearson test \cite{neyman1933problem} in regimes where explicit clutter likelihoods are not tractable.

\subsection{Weighted Log-\textit{p} Fusion with ANMF}
\label{ssec:fusion}

To exploit complementarity between the data-driven CVAE and the model-based ANMF, we map each detector score onto a common significance scale via the Probability Integral Transform (PIT) under $H_0$~\cite{Rosenblatt1952,Diebold1998,Gneiting2007}. 
For each Doppler bin $b$, we build Empirical Cumulative Distribution Functions (ECDFs) $\widehat{F}^{\text{CVAE}}_b$ and $\widehat{F}^{\text{ANMF}}_b$ from $H_0$ evaluation data and compute
\begin{equation*}
p_b^{\text{CVAE}} = 1-\widehat{F}^{\text{CVAE}}_b\!\big(s^{\text{CVAE}}\big), \hspace{0.2cm}
p_b^{\text{ANMF}} = 1-\widehat{F}^{\text{ANMF}}_b\!\big(s^{\text{ANMF}}\big)\,.
\end{equation*}
In the ideal case, where $\widehat{F}_b$ coincides with the true null Cumulative Distribution Functions (CDF), the PIT variables are exactly uniform on $(0,1)$ under $H_0$, so that
$p_b^{\text{CVAE}}$ and $p_b^{\text{ANMF}}$ are valid $p$-values, and any per-bin quantile thresholding yields an exact CFAR detector.
Using the empirical CDF $\widehat{F}_{H_0}$ built from $N_b$ background snapshots, the CFAR deviation is uniformly controlled at rate $O\left(N_b^{-1/2}\right)$ by the classical Dvoretzky-Kiefer-Wolfowitz concentration inequality for empirical CDFs \cite{dkw_massart1990}; a short derivation is given in the online supplementary material.\\

We then define the fusion score as a directed weighted Fisher/Lancaster combination ~\cite{Fisher1932,Lancaster1961}:
\begin{equation}
\label{eq:fused}
S_b^\star \;=\; -\left(w_b\log p_b^{\text{ANMF}}+(1-w_b)\log p_b^{\text{CVAE}}\right),\hspace{0.2cm} w_b\in[0,1]\, .
\end{equation}
where the weights \(w_b\) are chosen arbitrarily. In the following, we propose a strategy to recover CFAR by re-estimating the null CDF of $S_b^\star$ from clutter-only data.\\

\paragraph*{Choice of weights \(w_b\)}
We consider two practical options:  
(i) a data-driven schedule based on the CVAE’s null $p$-values;  
(ii) a prior shaping that emphasizes ANMF around low Doppler.

For the data-driven schedule, we define the weights $w_b$ as:
\begin{equation}
\label{eq:wb-sigmoid-final}
w_b \;=\; \frac{1}{1+\exp\!\left(-\displaystyle\frac{\bar p_b-\mu_p}{\sigma_p}\right)}\,,
\end{equation}
where $\bar p_b=\displaystyle\frac{1}{N_b}\sum_{i=1}^{N_b}p^{\text{CVAE}}_{b,i}$, $\mu_p=\displaystyle\frac{1}{m}\sum_{b=1}^{m}\bar p_b$, and where 
$\sigma_p = \displaystyle\frac{1}{m} \sum_{b=1}^m (\mu_p - p_b)^2 $.
 
This upweights ANMF in regions where the CVAE appears less discriminative under $H_0$, without requiring any prior on $b$.

When low Doppler is known to be challenging, we also consider a Gaussian prior centered at a reference bin $b_0$:
\begin{equation}
\label{eq:wb-prior}
w_b \;=\; \exp\!\left(-\displaystyle\frac{1}{2}\left(\displaystyle\frac{d(b,b_0)}{\sigma_0}\right)^2\right)\,,
\end{equation}
where $d(b,b_0)$ defines the distance between $b$ and $b_0$ Doppler bins and where $\sigma_0$ is the Doppler parameter width. In our experiments, we set $b_0$ to the zero-Doppler bin, which concentrates the most difficult clutter/target discrimination cases.\\

\paragraph*{Empirical CFAR calibration under dependence}
Since CVAE and ANMF may be dependent, we estimate the null law of $S_b^\star$ empirically using paired $H_0$ samples (same index for $p_b^{\text{CVAE}}$ and $p_b^{\text{ANMF}}$), and set bin-wise thresholds via a finite-sample quantile:
\begin{equation}
\lambda_{\text{fusion}}(b)
\;=\;
F^{-1}_{S_b^\star\,|\,H_0}\!\big(1-P_{\mathrm{fa}}\big)\,.
\end{equation}

\begin{algorithm}[!t]
\label{ssec:algo}
\caption{CVAE--ANMF Detection with Whitening and Log-$p$ Fusion}
\label{alg:cvae_detection}
\begin{algorithmic}[1]
\REQUIRE 
    Training set $\mathcal{D}_{\text{train}}$ (clutter, $H_0$), 
    evaluation set $\mathcal{D}_{\text{eval}}$ (clutter, $H_0$), 
    test set $\mathcal{D}_{\text{test}}$, 
    target false-alarm level $P_{\mathrm{fa}}$
\ENSURE 
    Detection decisions on $\mathcal{D}_{\text{test}}$ (per Doppler bin $b$)

\STATE \textbf{Preprocessing (pre-whitening before Doppler):}
       For each CPI in $\mathcal{D}_{\text{train}}$, $\mathcal{D}_{\text{eval}}$, and $\mathcal{D}_{\text{test}}$,
       apply the segmentation and local whitening of Alg.~\ref{alg:whitening}:
       \begin{itemize}
         \item extract slow-time snapshots $\mathbf{y}_{r,p}\in\mathbb{C}^m$,
         \item estimate a local covariance $\hat{\mathbf{R}}_{\mathcal{N}_r}$ from adjacent ranges
               and form a whitening operator $\hat{\mathbf{R}}_{\mathrm{reg}}^{-1/2}$,
         \item whiten $\mathbf{y}_{r,p}^{\mathrm{w}} = \hat{\mathbf{R}}_{\mathrm{reg}}^{-1/2}\,\mathbf{y}_{r,p}$,
         \item compute the Doppler profiles 
               $\mathbf{z}_{r,p}^{\mathrm{w}} = \mathrm{DFT}\!\big(\mathbf{y}_{r,p}^{\mathrm{w}}\big)$.
       \end{itemize}
       Denote by $\mathcal{Z}_{\text{train}}$, $\mathcal{Z}_{\text{eval}}$, and $\mathcal{Z}_{\text{test}}$
       the resulting collections of (optionally) whitened Doppler profiles 
       $\mathbf{z}_{r,p}^{\mathrm{w}}$.
       For notational simplicity, we write $\mathbf{z}$ for such profiles in the sequel.

\STATE \textbf{CVAE training:}
       Train the complex VAE on $\mathcal{Z}_{\text{train}}$ under $H_0$
       using the $\beta$-ELBO objective in~\eqref{eq:cvae-elbo}.

\STATE \textbf{Score computation on $\mathcal{Z}_{\text{eval}}$:}
       For each Doppler bin $b$ and each evaluation profile $\mathbf{z}\in\mathcal{Z}_{\text{eval}}$,
       compute the CVAE reconstruction score $s^{\mathrm{CVAE}}(\mathbf{z})$
       as in~\eqref{eq:score} and the ANMF statistic
       $s^{\mathrm{ANMF}}(\mathbf{z})$ (Sec.~\ref{sec:statmodel}).

\STATE \textbf{Null calibration and fusion thresholds:}
       For each bin $b$:
       \begin{enumerate}
           \item Build empirical CDFs $\widehat{F}^{\mathrm{CVAE}}_b$ and
                 $\widehat{F}^{\mathrm{ANMF}}_b$ from the $H_0$ scores
                 on $\mathcal{Z}_{\text{eval}}$.
           \item Map scores to $p$-values via the PIT:
                 $p^{\mathrm{CVAE}}_b = 1-\widehat{F}^{\mathrm{CVAE}}_b(s^{\mathrm{CVAE}})$,
                 $p^{\mathrm{ANMF}}_b = 1-\widehat{F}^{\mathrm{ANMF}}_b(s^{\mathrm{ANMF}})$.
                 
           \item Choose weights $w_b$ (e.g., sigmoid schedule or Gaussian prior)
                 and form the fused statistic $S_b^\star$ as in~\eqref{eq:fused}
                 from paired $p$-values
                 $(p^{\mathrm{CVAE}}_b,p^{\mathrm{ANMF}}_b)$.
           \item Estimate the null CDF $F_{S_b^\star|H_0,b}$ from the fused
                 $H_0$ samples and set the fusion threshold
                 $\lambda_{\mathrm{fusion}}(b)
                 = F^{-1}_{S_b^\star|H_0,b}(1-P_{\mathrm{fa}})$.
       \end{enumerate}

\STATE \textbf{Testing on $\mathcal{Z}_{\text{test}}$:}
       For each test profile $\mathbf{z}\in\mathcal{Z}_{\text{test}}$ and each
       Doppler bin $b$:
       \begin{enumerate}
           \item Compute $s^{\mathrm{CVAE}}(\mathbf{z})$ and
                 $s^{\mathrm{ANMF}}(\mathbf{z})$.
           \item Map to $p$-values using the precomputed CDFs:
                 $p^{\mathrm{CVAE}}_b$, $p^{\mathrm{ANMF}}_b$.
           \item Form $S_b^\star$ via~\eqref{eq:fused} and decide
                 \[
                   \text{declare } H_1 \text{ at bin } b
                   \iff S_b^\star \ge \lambda_{\mathrm{fusion}}(b).
                 \]
       \end{enumerate}
\RETURN Decisions $\{H_0,H_1\}$ on all Doppler bins of $\mathcal{Z}_{\text{test}}$.
\end{algorithmic}
\end{algorithm}

\begin{figure*}[!t]
\centering
\resizebox{0.97\textwidth}{!}{%
\begin{tikzpicture}[
    font=\large,
    >=Stealth,
    node distance=10mm and 18mm,
    every node/.style={transform shape},
    box/.style={
        draw,
        rounded corners=2.5mm,
        line width=1pt,
        align=center,
        inner sep=6pt,
        minimum height=10mm,
        fill=#1!10,
        text width=4.6cm
    },
    smallbox/.style={
        draw,
        rounded corners=2.0mm,
        line width=0.9pt,
        align=center,
        inner sep=4pt,
        minimum height=8mm,
        fill=#1!5,
        text width=4.4cm
    },
    flow/.style={-Stealth, line width=1.2pt, line cap=round},
    optflow/.style={-Stealth, line width=1.0pt, line cap=round, dashed},
    branch/.style={line width=1.2pt, line cap=round},
    branchopt/.style={line width=1.0pt, line cap=round, dashed},
    lab/.style={font=\normalsize\bfseries}
]

\definecolor{cIn}{RGB}{52,58,64}
\definecolor{cSeg}{RGB}{79,98,214}
\definecolor{cWhit}{RGB}{243,125,16}
\definecolor{cCVAE}{RGB}{13,110,253}
\definecolor{cANMF}{RGB}{111,66,193}
\definecolor{cPIT}{RGB}{214,51,132}
\definecolor{cFuse}{RGB}{25,135,84}

\node[box=cIn] (z)
  {$\mathbf{Y} \in \mathbb{C}^{N_{\text{ranges}}\times N_{\text{pulses}}}$};

\node[box=cSeg, above= of z] (seg)
  {Profiles per range gate\\[1mm]
   $\mathbf{y}_{r,p}\in\mathbb{C}^{N_{\text{pulse}}}$};

\node[box=cWhit, right=of seg] (white)
  {Local whitening (optional)\\[1mm]
   $\mathbf{y}^\mathrm{w}_{r,p}=\hat{\mathbf{R}}_\mathrm{reg}^{-\frac{1}{2}} \mathbf{y}_{r,p}$\\[0.5mm]
   
   };

\node[box=cIn, below=5mm of white] (dft)
  {Discrete Fourier transform\\[1mm]
   $\mathbf{z} = \mathrm{DFT}\left[\mathbf{y}^\mathrm{w}_{r,p}\right]$\\[0.5mm]
   };

\draw[flow] (z) -- (seg) node[midway, above=-10pt] {Preprocessing};
\draw[flow] (seg) -- (white);
\draw[flow] (white) -- (dft);

\coordinate (branchRaw) at ($(seg.east)+(10mm,0)$);

\node[box=cCVAE, right=15mm of white] (cvae)
  {Encoder($\mathbf{z}$) $\rightarrow (\boldsymbol{\mu},\boldsymbol{\sigma}^{2},\boldsymbol{\delta})$\\[0.5mm]
   complex reparam. $\rightarrow \mathbf{x}$\\[0.5mm]
   Decoder $\rightarrow \hat{\mathbf{z}}$};

\node[lab, above=9mm of cvae, text=cCVAE!90!black]
  {Complex VAE (CVAE)};

\node[smallbox=cCVAE, below=5mm of cvae] (statc)
  {$s^{\text{CVAE}} = \displaystyle\sum_n |z_n-\hat{z}_n|^2$};

\node[smallbox=cPIT, below=of statc] (pitc)
  {per Doppler bin $b$\\[0.5mm]
   $p^{\text{CVAE}}_b = \widehat{F}^{\text{CVAE}}_b(s)$};

\node[lab, above =3mm of pitc.north, text=cPIT!90!black]
 { PIT / ECDF under $H_0$ \hspace{3mm} };

\node[lab, rotate=90, below right=4mm of cvae, text=cCVAE!90!black]
  {Data-driven};

\node[box=cANMF, right=15mm of cvae] (anmf)
  {$\displaystyle
    \Lambda_{\text{ANMF}} =
    \frac{|\mathbf{p}^{H}\widehat{\boldsymbol{\Sigma}}^{-1}\mathbf{z}|^{2}}
         {(\mathbf{p}^{H}\widehat{\boldsymbol{\Sigma}}^{-1}\mathbf{p})
          (\mathbf{z}^{H}\widehat{\boldsymbol{\Sigma}}^{-1}\mathbf{z})}$\\[1mm]
   $\widehat{\boldsymbol{\Sigma}}$: SCM or Tyler FP};

\node[lab, above=6mm of anmf, text=cANMF!90!black]
  {Adaptive Normalized Matched Filter (ANMF)};

\node[smallbox=cANMF, below=5mm of anmf] (stata)
  {$s^{\text{ANMF}} = \Lambda_{\text{ANMF}}$};

\node[smallbox=cPIT, below=of stata] (pita)
  {per Doppler bin $b$\\[0.5mm]
   $p^{\text{ANMF}}_b = \widehat{F}^{\text{ANMF}}_b(s)$};

\node[lab, above=3mm of pita.north, text=cPIT!90!black]
  {PIT / ECDF under $H_0$ \hspace{3mm} };

\node[lab, rotate=90, below right=4mm of anmf, text=cANMF!90!black]
  {Model-based};

\coordinate (underCVAE)   at ($(cvae.south)+(0,-10mm)$);
\coordinate (overCVAERaw) at ($(cvae.north)+(0,3mm)$);

\coordinate (branchPre) at ($(white.north east)+(5mm,9mm)$);

\draw[flow] (dft.east) -| (branchPre) -| (cvae.north);
\draw[flow] (dft.east) -| (branchPre) -| (anmf.north);


\draw[flow] (cvae.south) -- (statc.north);
\draw[flow] (statc.south) -- (pitc.north);

\draw[flow] (anmf.south) -- (stata.north);
\draw[flow] (stata.south) -- (pita.north);

\coordinate (midPIT) at ($(pitc.south)!0.5!(pita.south)$);

\node[box=cFuse] (fusion)
  at ($(pitc)+(-65mm,-9mm)$)
  {$S_b^\star
    = -\big[w_b \log p_b^{\text{ANMF}}
           + (1-w_b)\log p_b^{\text{CVAE}}\big]$\\[1mm]
   $w_b \in [0,1]$ (e.g., Gaussian vs.\ Doppler-dependent weight)};

\node[lab, above=2mm of fusion, text=cFuse!90!black]
  {Weighted log-$p$ fusion};

\node[box=cFuse, left=of fusion] (cfar)
  {$\lambda_{\text{fusion}}(b)
      = F^{-1}_{S_b^\star|H_0,b}(1-P_{\mathrm{fa}})$\\[1mm]
   Decision per Doppler bin $b$: $H_1$ vs.\ $H_0$};

\node[lab, above=1mm of cfar.north, text=cFuse!90!black]
  {CFAR threshold};

\draw[flow] (pitc.south)  |- ($(fusion.east) + (0,0mm)$);
\draw[flow] (pita.south)  |- ($(fusion.east) + (0,-4mm)$);

\draw[flow] (fusion) -- (cfar);

\end{tikzpicture}%
}
\caption{Two-branch detection pipeline combining data-driven and model-based approaches.}
\label{fig:cvae-anmf-fusion}
\end{figure*}
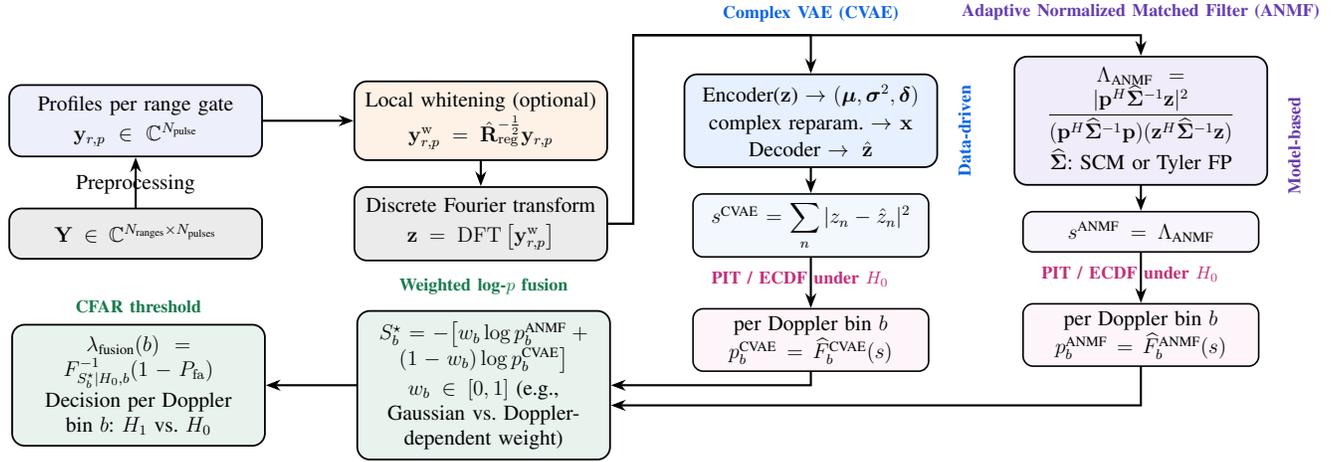

\section{Results and Simulations}
\label{sec:results}

We evaluate the proposed CVAE-based detector and its decision-level fusion with ANMF (Section~\ref{ssec:fusion}) against MF, NMF, AMF-SCM, and ANMF-Tyler on both simulated noise environments and real sea clutter. In all experiments, we report the probability of detection $P_d$ as a function of the signal-to-noise ratio (SNR) at a fixed false-alarm probability $P_{fa}=10^{-2}$, and we provide results both before and after local whitening (Section~\ref{ssec:datawhitening}). 

\subsection{Experimental Setups}
\label{ssec:exp_setups}

\subsubsection{Signal and noise characteristics} 
\label{ssec:signoise}

The complex target echo is modeled as
\begin{equation}
  \alpha = \sqrt{\frac{\mathrm{SNR}}{m}} \, e^{2j\pi \phi}, 
  \qquad \phi \sim \mathcal{U}([0,1])\,,
\end{equation}
with Doppler steering vector
\begin{equation}
  \mathbf{p} = \left(1, e^{2j\pi d/m}, \ldots, e^{2j\pi d(m-1)/m} \right)^T\,,
\end{equation}
for $m = 16$ Doppler bins, where $d\in \{0, \ldots, m-1\}$ denotes the $(d{+}1)$th bin.

The disturbance environments considered in the simulation are:

\begin{itemize}
\item \textbf{Correlated Gaussian Noise (cGN)}:\\
$\mathbf{c}\sim\mathcal{CN}(\mathbf{0}, \boldsymbol{\Sigma}_c)$, where $\boldsymbol{\Sigma}_c = \boldsymbol{\mathcal{T}}(\rho)$ is a Toeplitz matrix with correlation coefficient $\rho = 0.5$.

\item \textbf{Correlated Compound Gaussian Noise (cCGN)}:\\
$\mathbf{z} = \sqrt{\tau}\,\mathbf{c}$, with $\mathbf{c} \sim \mathcal{CN}(\mathbf{0}, \boldsymbol{\Sigma}_c)$ and $\tau$ a Gamma-distributed texture, $\tau \sim \Gamma(\mu, 1/\mu)$ with $\mu=1$, representing heavy-tailed correlated clutter.

\item \textbf{Additive White Gaussian Noise (AWGN)}:\\
$\mathbf{n} \sim \mathcal{CN}(\mathbf{0}, \sigma_n^2 \,\mathbf{I})$, modelling thermal noise with power $\sigma_n^2$.

\end{itemize}

Then, we consider mixed environments where the thermal noise is added to the clutter component. In the Gaussian Noise case (cGN+AWGN), the disturbance is given by $\mathbf{z} = \mathbf{c} + \mathbf{n}$, while in the Compound Gaussian case (cCGN+AWGN) we use $\mathbf{z} = \sqrt{\tau}\,\mathbf{c} + \mathbf{n}$.

For adaptive detectors, the covariance matrix is estimated using the SCM or Tyler’s estimator with $K=2m$ secondary data. 


\subsubsection{CSIR sea clutter data}
\label{ssec:csirdataset}

We validate the proposed approach on two CSIR maritime X-band recordings (South African coast). For concision, we retain the fields most relevant to detection and reproducibility. Table~\ref{tab:csir_meta} summarizes radar and processing metadata.

\begin{table}[!t]
\centering
\caption{CSIR metadata (instrument/radar/processing).}
\label{tab:csir_meta}
\footnotesize
\setlength{\tabcolsep}{5pt}
\begin{tabular}{lcc}
\hline
 & \textbf{CFA16-002} & \textbf{CFA16-008}\\
\hline
Type & Sea clutter & Sea clutter\\
Transmitter frequency & 6.9\,GHz & 6.9\,GHz\\
Pulse Repetition Frequency & 5\,kHz & 5\,kHz\\
Range resolution &  15\,m &  15\,m  \\
Range extent & 1440\,m (96 gates) & 1440\,m (96 gates)\\
\hline
\end{tabular}
\end{table}


\begin{figure}[!t]
\centering
\subfloat[CFA16-002: range--time map.]{%
  \includegraphics[width=0.23\textwidth,trim={48mm 0 40mm 0},clip]{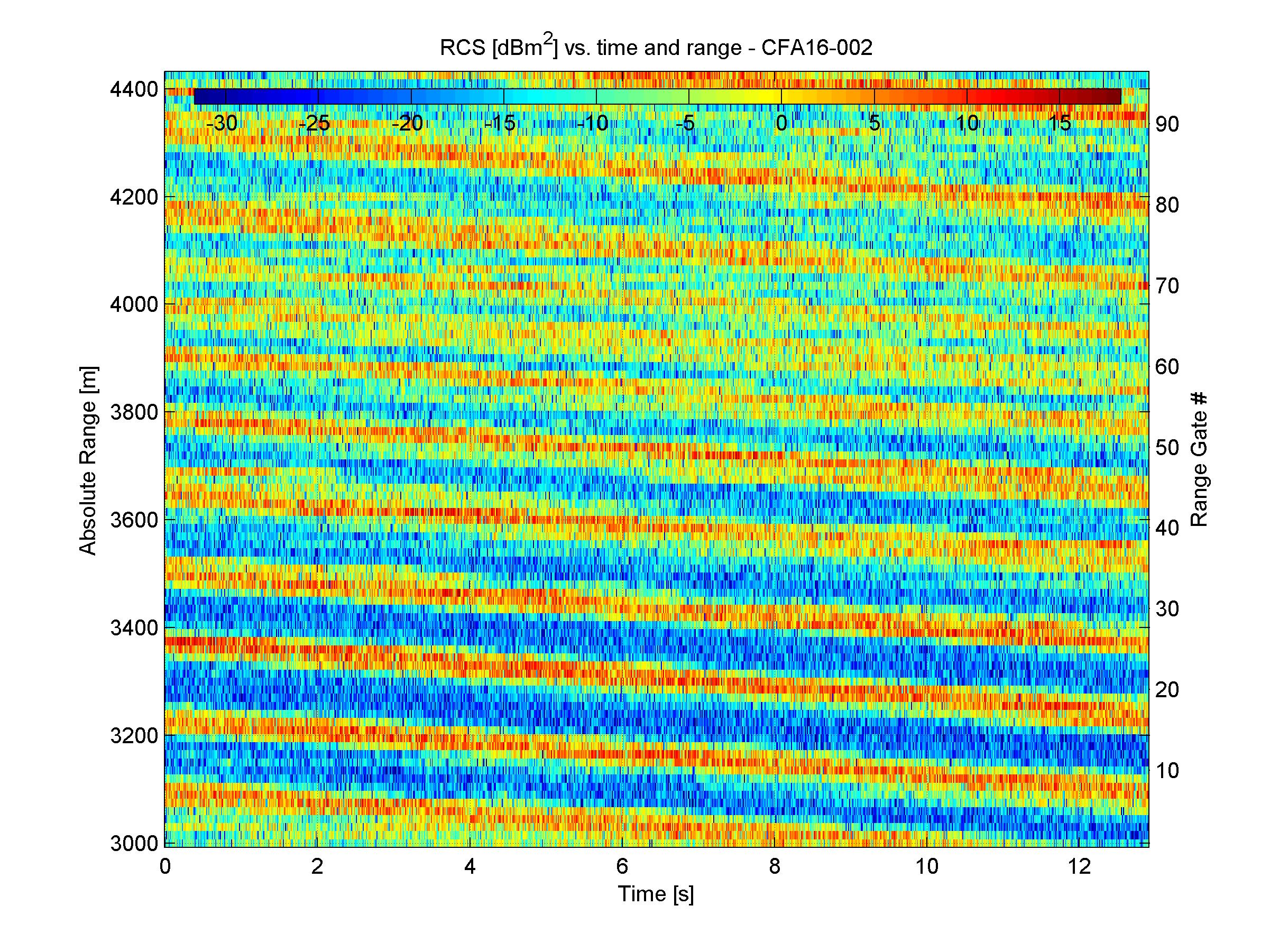}%
  \label{fig:cfa002_rt}
}\hfill
\subfloat[CFA16-002: Doppler--time (gate 54).]{%
  \includegraphics[width=0.23\textwidth,trim={40mm 0 60mm 0},clip]{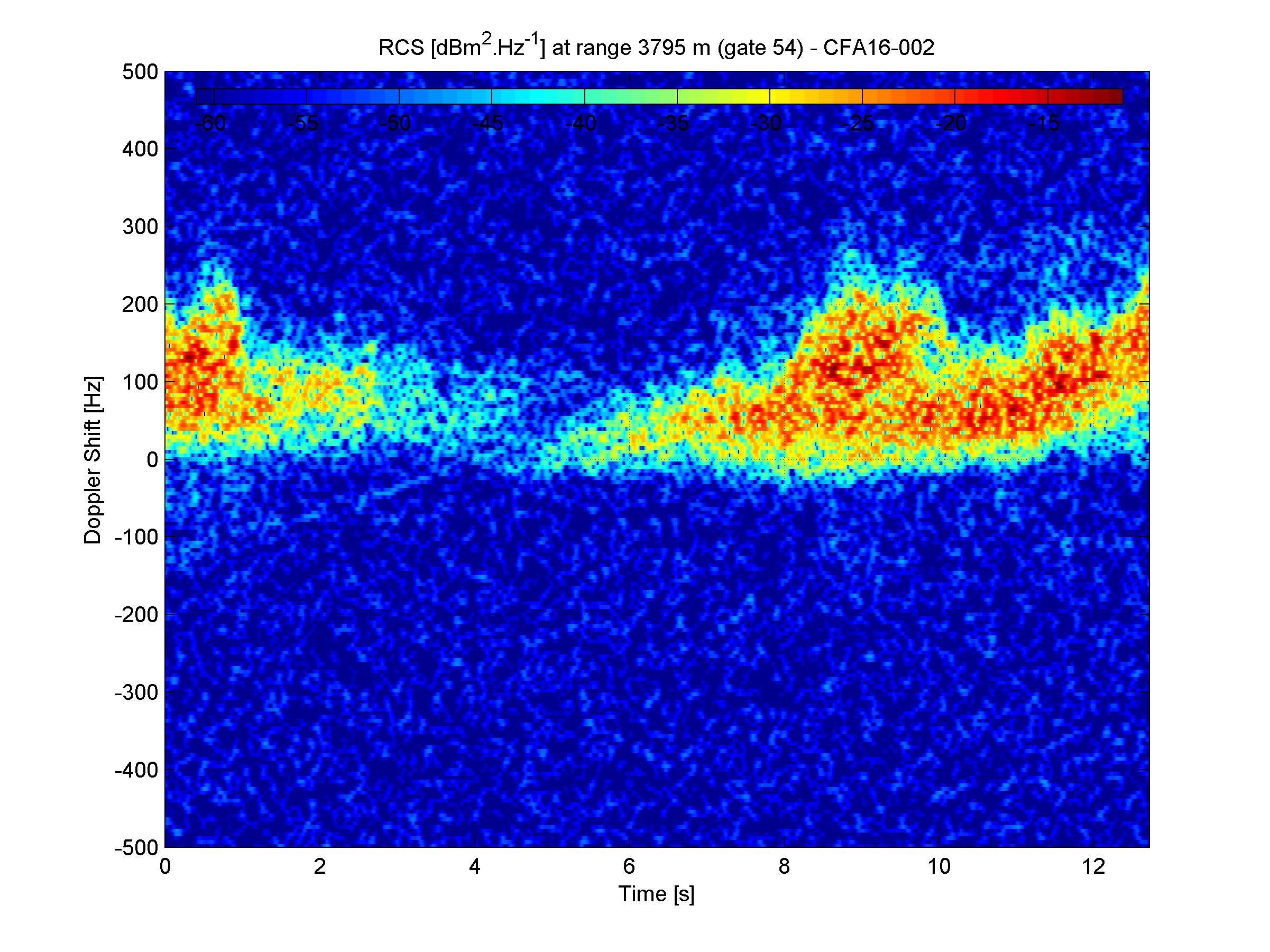}%
  \label{fig:cfa002_dt}
}\\[2mm]
\subfloat[CFA16-008: range--time map.]{%
  \includegraphics[width=0.23\textwidth,trim={48mm 0 40mm 0},clip]{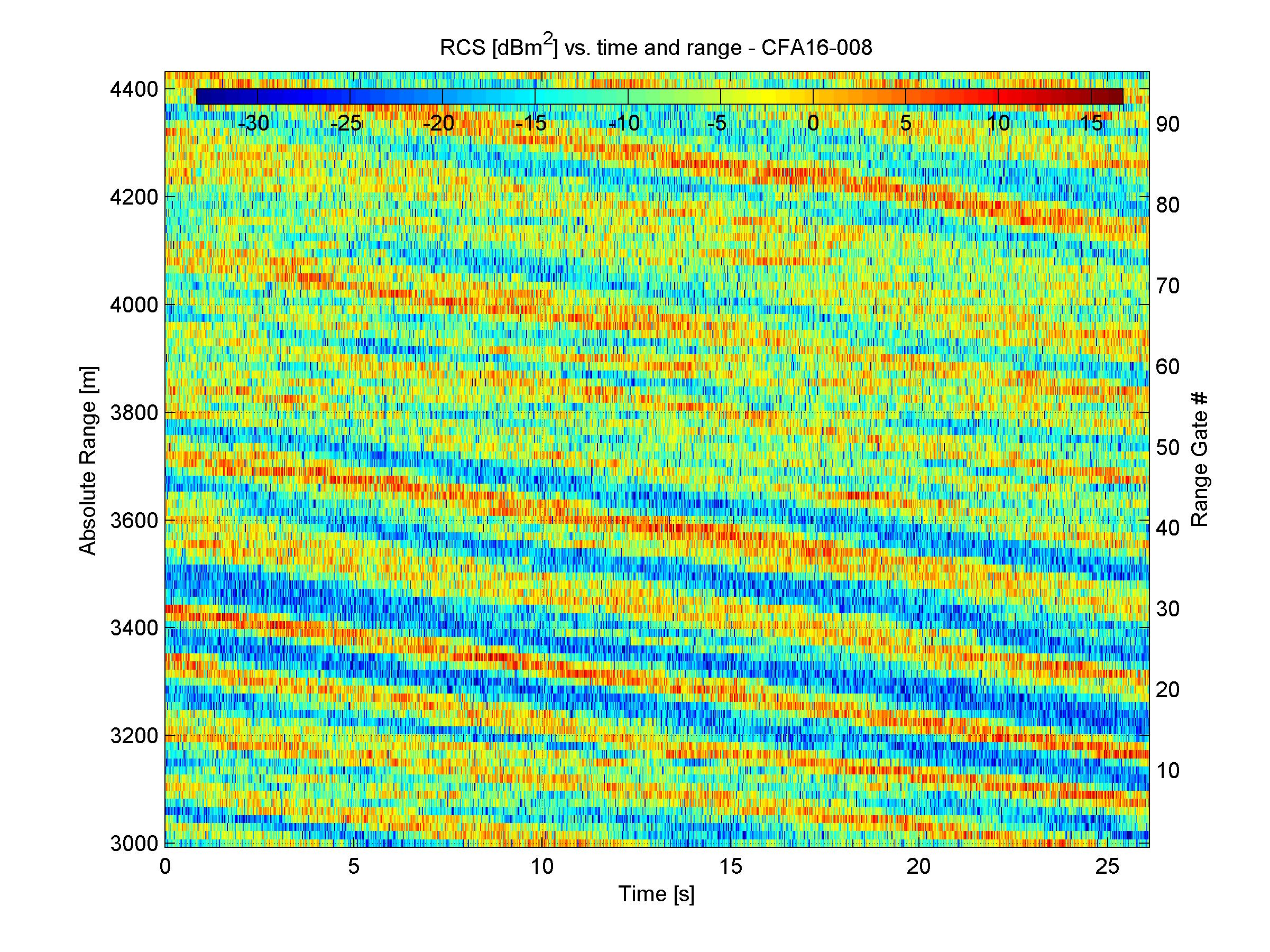}%
  \label{fig:cfa008_rt}
}\hfill
\subfloat[CFA16-008: Doppler--time (gate 40).]{%
  \includegraphics[width=0.23\textwidth,trim={40mm 0 60mm 0},clip]{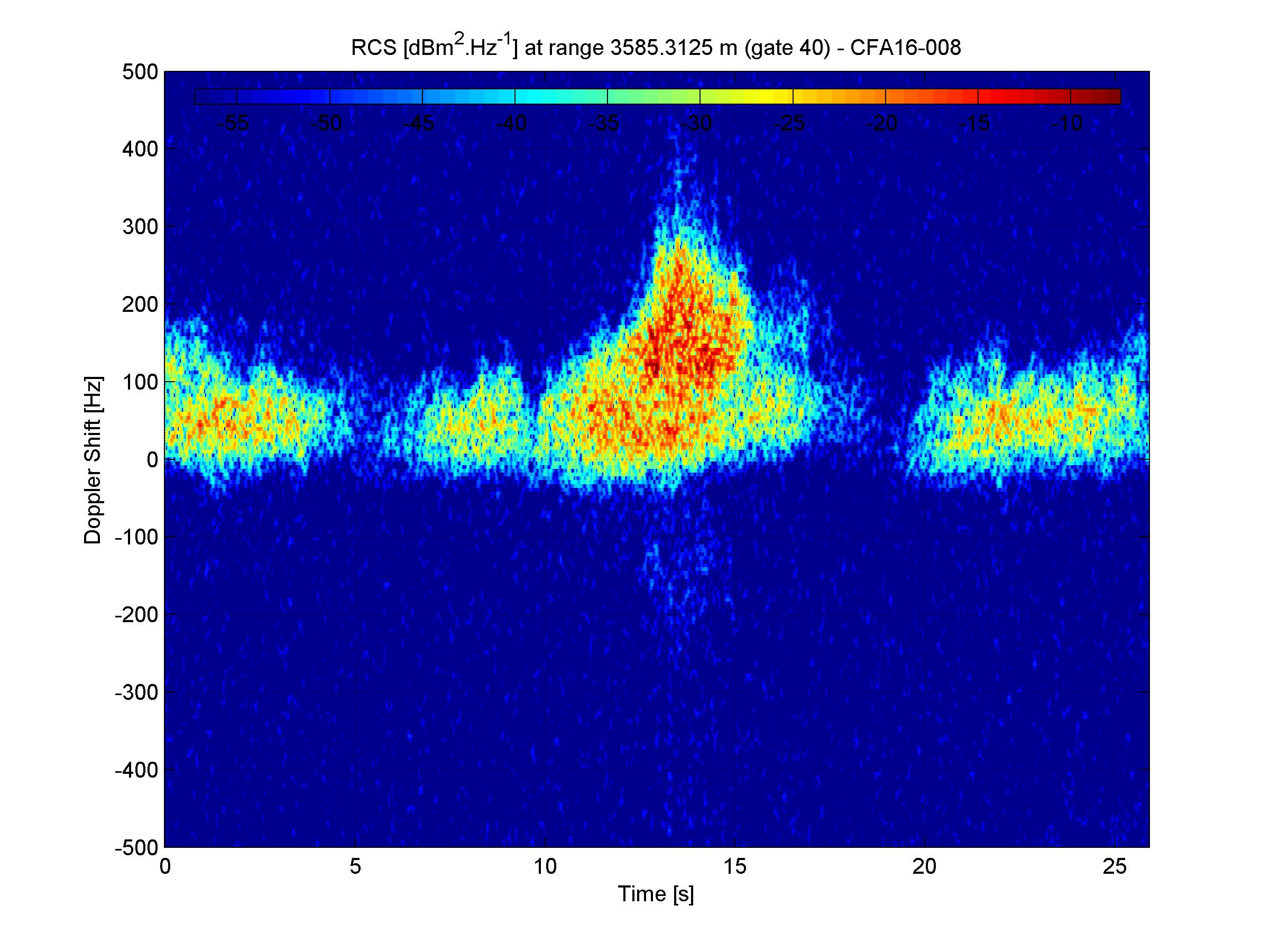}%
  \label{fig:cfa008_dt}
}
\caption{CSIR sea-clutter datasets used in Section~\ref{sec:results}. 
Left: range-time maps (96 range gates, 15\,m resolution). 
Right: Doppler-time spectrograms at the gates used for detection analysis. 
Same color scale across panels.}
\label{fig:csir_overview}
\end{figure}

\subsection{Simulated Data: Zero-Doppler ($d=0$) Curves}
\label{ssec:zerodoppler_simu}

\begin{figure*}[!t]
\centering
\subfloat[cGN+AWGN.]{%
  \includegraphics[width=0.32\textwidth,trim={5mm 3mm 16mm 12mm},clip]{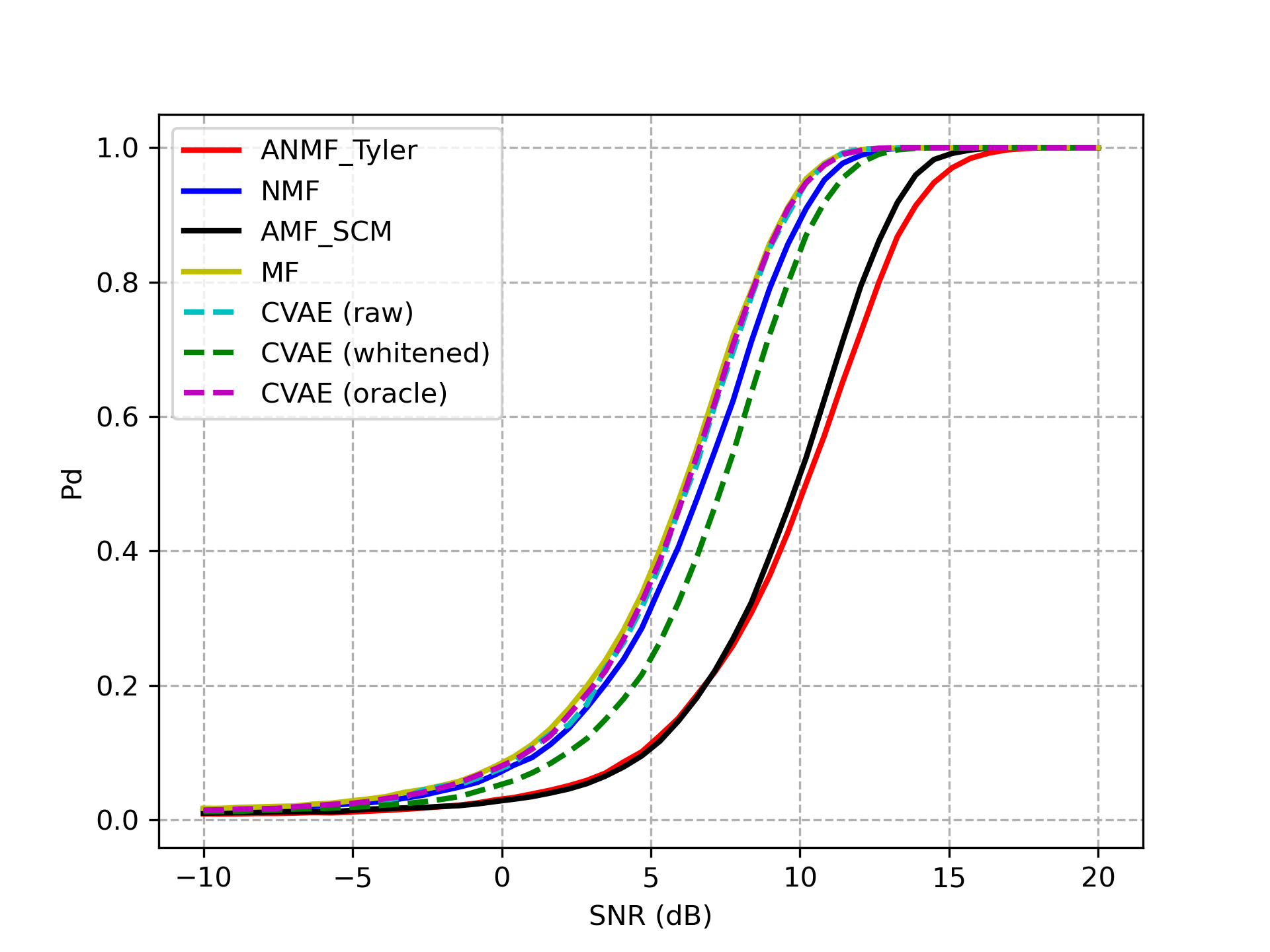}%
  \label{fig:pd_snr_cgn_awgn}
}\hfil
\subfloat[cCGN.]{%
  \includegraphics[width=0.32\textwidth,trim={5mm 3mm 16mm 12mm},clip]{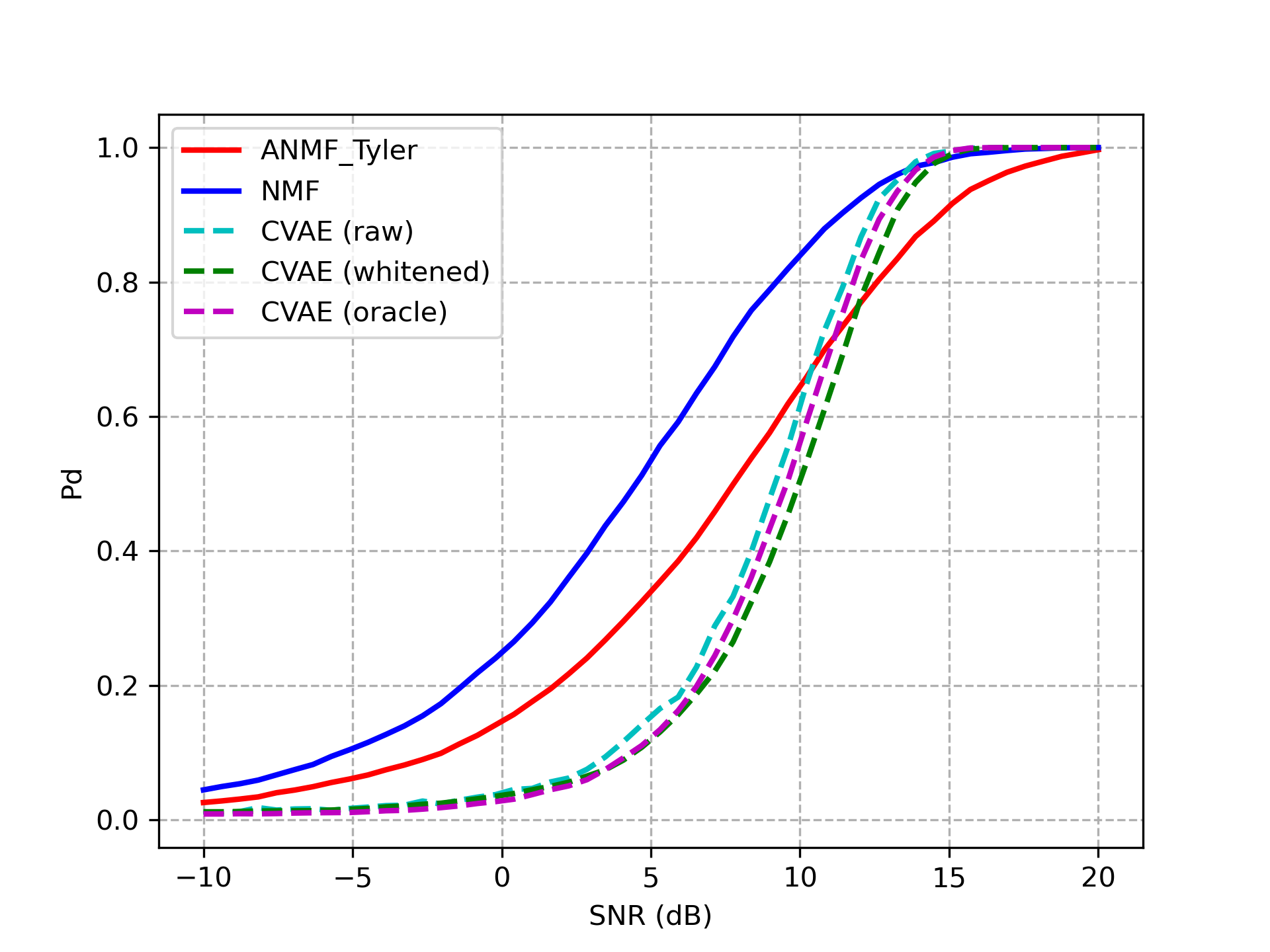}%
  \label{fig:pd_snr_ccgn}
}\hfil
\subfloat[cCGN+AWGN.]{%
  \includegraphics[width=0.32\textwidth,trim={5mm 3mm 16mm 12mm},clip]{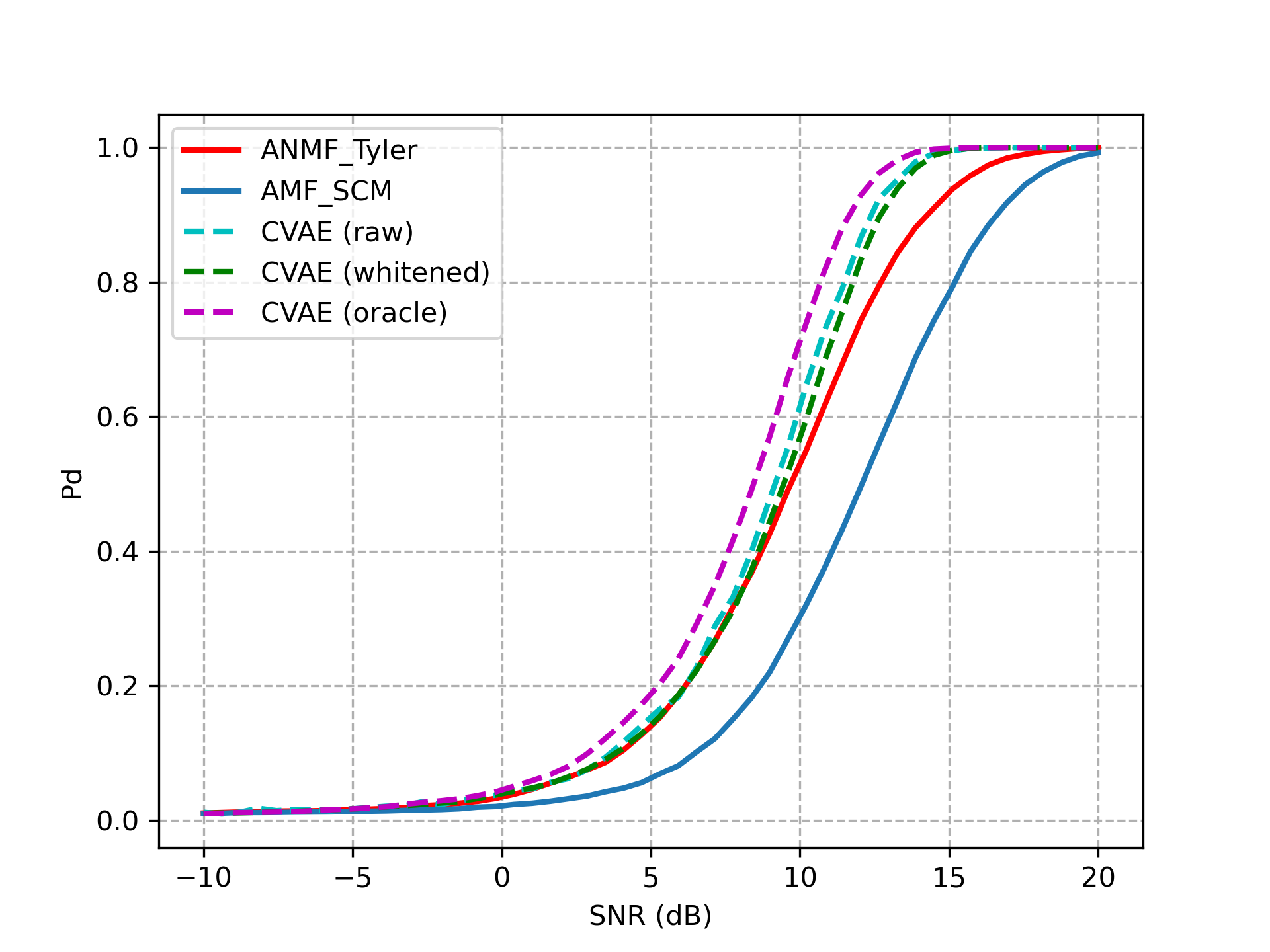}%
  \label{fig:pd_snr_ccgn_awgn}
}
\caption{$P_d$ vs.\ SNR at $d=0$ under different simulated noise models 
($P_{fa}=10^{-2}$, $\rho=0.5$, $\mu=1$, $m=16$, $K=32$).}
\label{fig:cpdsnr}
\end{figure*}

Figure~\ref{fig:cpdsnr} reports $P_d$ vs.\ SNR at Doppler bin $d=0$ under cGN+AWGN, cCGN, and cCGN+AWGN. Each panel compares MF/NMF baselines, AMF-SCM or ANMF-Tyler, and the CVAE detector (raw, locally whitened, and oracle-whitened when applicable). Oracle-whitened refers to the same process as local whitening, but uses the true covariance matrix, which is only available in simulation.

In the nearly Gaussian cGN+AWGN case (Fig.~\ref{fig:cpdsnr}(a)), the MF curve is an upper bound and both the raw and oracle-whitened CVAE closely match it, while NMF and locally whitened CVAE incur a small rightward shift and AMF-SCM/ANMF-Tyler show the largest SNR loss from covariance estimation. Under heavy-tailed cCGN (Fig.~\ref{fig:cpdsnr}(b)), NMF remains a strong benchmark but ANMF-Tyler degrades, whereas all CVAE variants yield steeper $P_d$--SNR transitions and approach NMF at medium-to-high SNR. When white noise is added (cCGN+AWGN, Fig.~\ref{fig:cpdsnr}(c)), all detectors benefit from partial ``Gaussianization'', but the CVAE achieves a given $P_d$ at noticeably lower SNR than ANMF-Tyler; oracle whitening gives a slight additional gain, whereas local whitening has limited impact. These trends are summarized in Table~\ref{tab:simu_summary}.

\subsection{Simulated Data: Full Doppler $P_d$–SNR Maps}
\label{ssec:alldoppler_simu}

We now examine $P_d$ as a joint function of SNR and Doppler bin for the simulated clutter models. Each heatmap shows AMF-SCM, ANMF-Tyler, and the CVAE variants over all $m=16$ Doppler bins.

\begin{figure*}[!t]
\centering
\includegraphics[width=\textwidth]{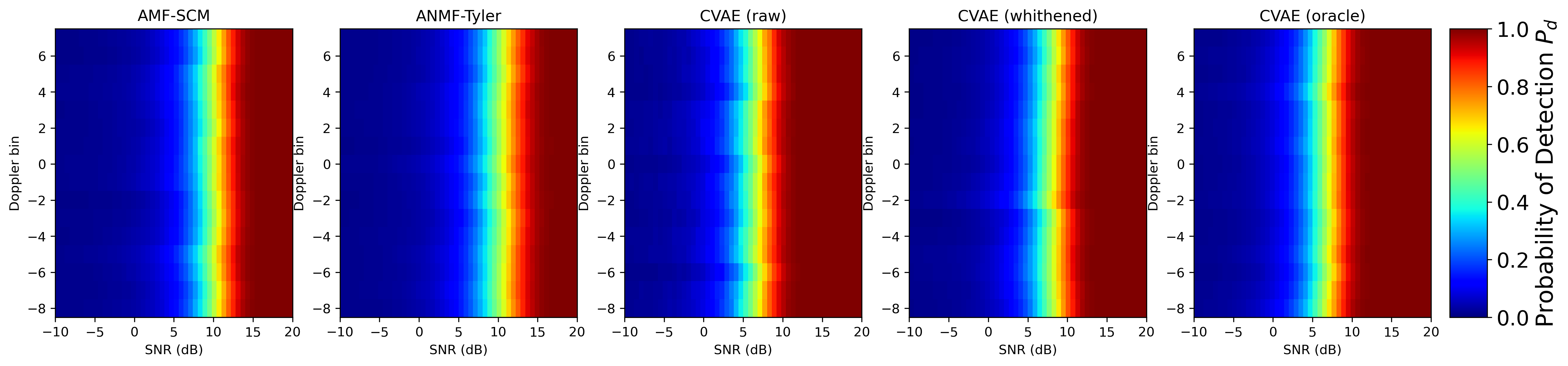}
\caption{$P_d$-SNR-Doppler maps for the cGN+AWGN setting ($P_{fa}=10^{-2}$). 
Left to right: AMF-SCM, ANMF-Tyler, CVAE (raw), CVAE (whitened), and oracle CVAE whitening.}
\label{fig:maps_cGN_AWGN}
\end{figure*}

\begin{figure*}[!t]
\centering
\includegraphics[width=\textwidth]{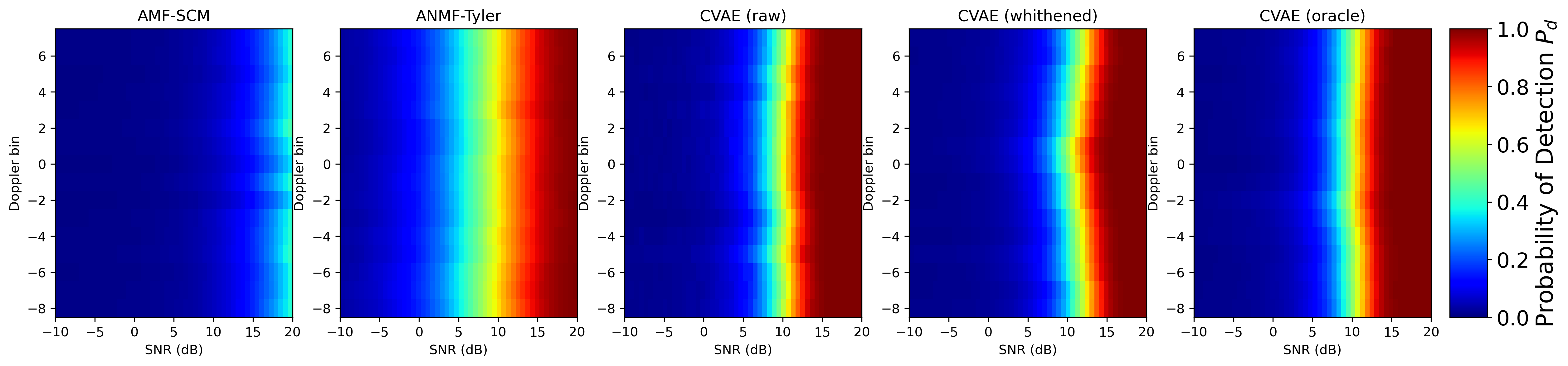}
\caption{$P_d$-SNR-Doppler maps for the cCGN setting (compound clutter, no white noise).}
\label{fig:maps_cCGN}
\end{figure*}

\begin{figure*}[!t]
\centering
\includegraphics[width=\textwidth]{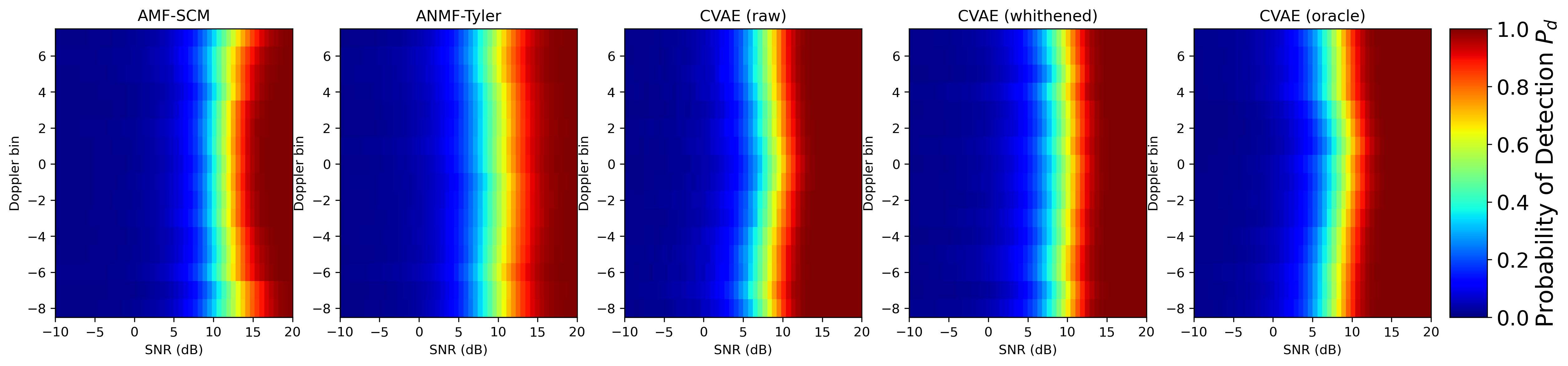}
\caption{$P_d$-SNR-Doppler maps for the cCGN+AWGN setting.}
\label{fig:maps_cCGN_AWGN}
\end{figure*}

For cGN+AWGN (Fig.~\ref{fig:maps_cGN_AWGN}), all detectors exhibit nearly vertical detection fronts, confirming Doppler-invariant homogeneous clutter once correlation is handled; the oracle CVAE aligns with the MF/NMF benchmark, and the raw CVAE reproduces this behavior, while local whitening and adaptive covariance estimation shift the fronts slightly to higher SNR. In cCGN (Fig.~\ref{fig:maps_cCGN}), AMF-SCM is clearly degraded and Doppler-dependent, ANMF-Tyler partly restores robustness but still requires several extra dB, and the raw/oracle CVAE yield the leftmost and sharpest fronts, nearly Doppler-invariant. Adding white noise (cCGN+AWGN, Fig.~\ref{fig:maps_cCGN_AWGN}) moves all fronts left and reduces Doppler dependence; AMF-SCM remains suboptimal, ANMF-Tyler stays behind the CVAE by roughly $2$--$3$\,dB over most bins, and local whitening has a limited smoothing effect. These qualitative rankings are consolidated in Table~\ref{tab:simu_summary}.

\begin{table*}[!t]
\centering
\caption{Qualitative ranking of detectors on simulated clutter
($P_{fa}=10^{-2}$). Symbols: ``++'' best or near-optimal, ``+'' good,
``0'' moderate, ``--'' clearly suboptimal in terms of $P_d$--SNR
(consistent with both the $d{=}0$ curves and the full Doppler maps).}
\label{tab:simu_summary}
\begin{tabular}{lcccccc}
\toprule
Scenario 
& MF/NMF 
& AMF-SCM 
& ANMF-Tyler 
& CVAE (raw) 
& CVAE (loc.\ white) 
& CVAE (oracle) \\
\midrule
cGN+AWGN      
& ++  
& 0   
& 0   
& ++  
& +   
& ++  
\\
cCGN          
& ++  
& --  
& 0   
& +   
& +   
& +   
\\
cCGN+AWGN     
&    
& --  
& 0   
& ++  
& +   
& ++  
\\
\bottomrule
\end{tabular}
\end{table*}

\subsection{CSIR Data: Zero-Doppler ($d=0$) Curves}
\label{ssec:zerodoppler_csir}

For the CSIR sea clutter, we report $P_d$ vs.\ SNR at $d=0$ for both scenes (CFA16-002 and CFA16-008) and for both raw and locally whitened preprocessing. The same CFAR calibration ($P_{fa}=10^{-2}$) is applied to all detectors.

\begin{figure*}[!t]
\centering
\subfloat[CFA16-002 (raw).]{%
  \includegraphics[width=0.4\textwidth]{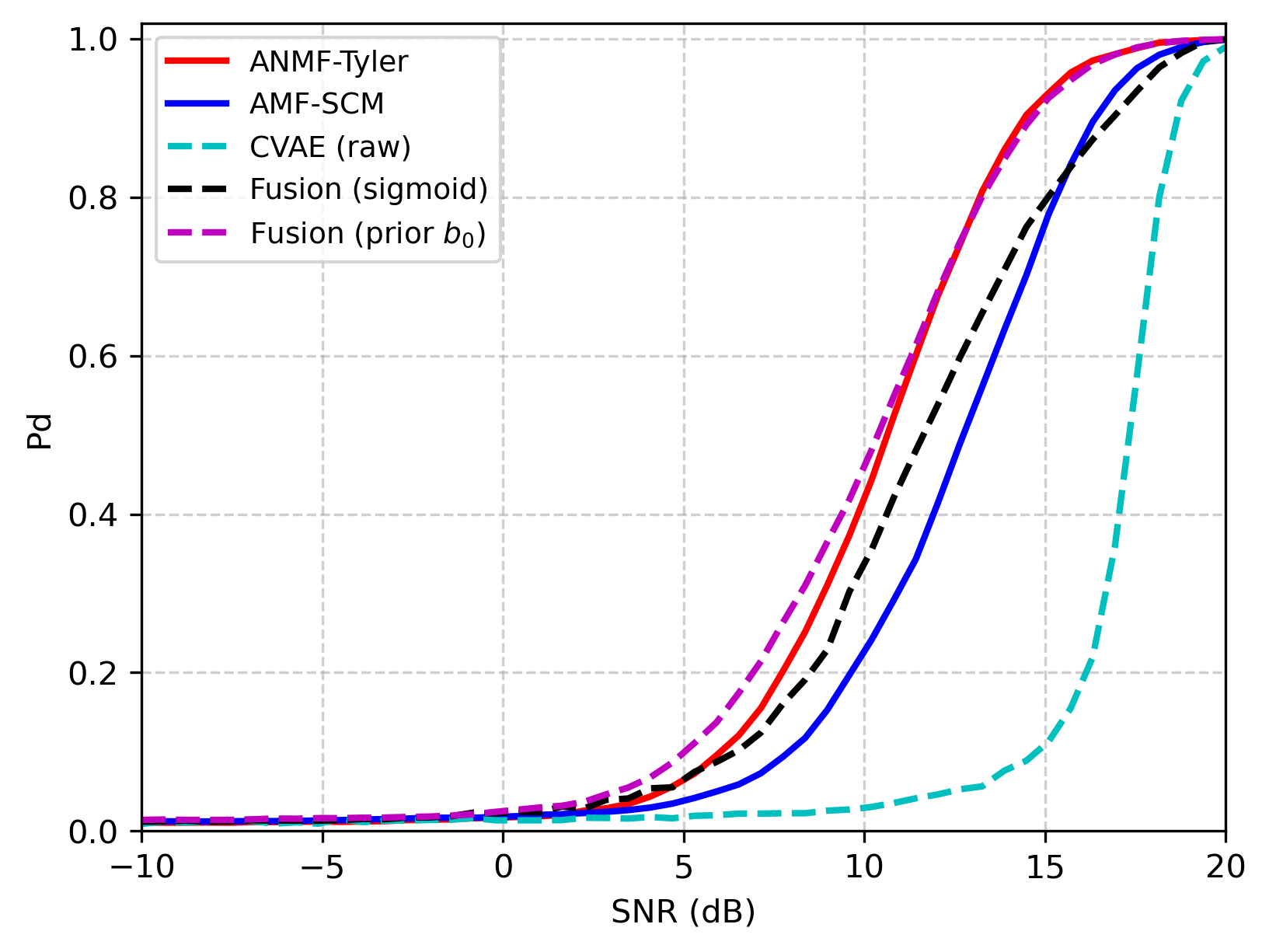}%
  \label{fig:pd_snr_002_raw}
}
\subfloat[CFA16-002 (whitened).]{%
  \includegraphics[width=0.4\textwidth]{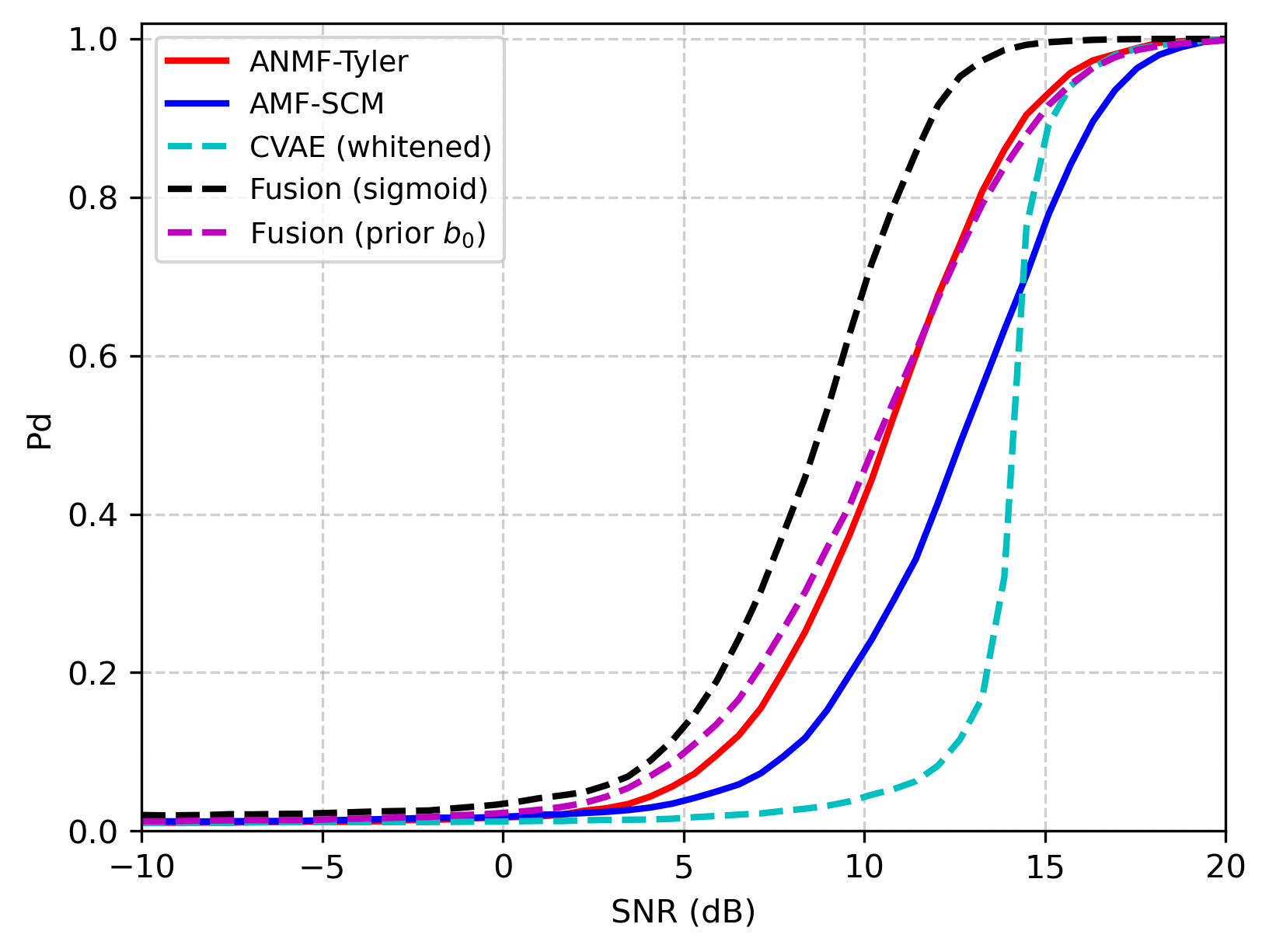}%
  \label{fig:pd_snr_002_white}
}\\
\subfloat[CFA16-008 (raw).]{%
  \includegraphics[width=0.4\textwidth]{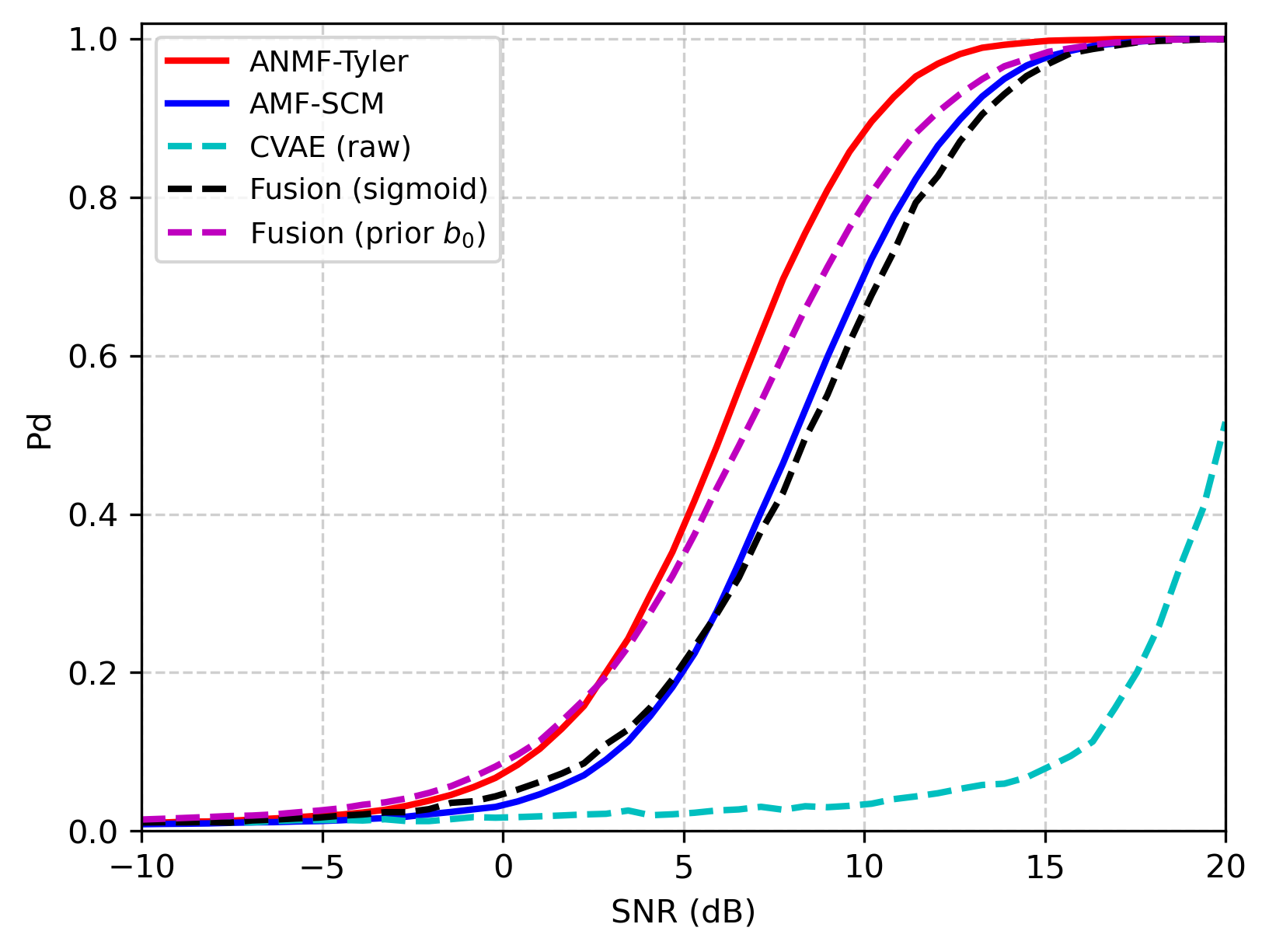}%
  \label{fig:pd_snr_008_raw}
}
\subfloat[CFA16-008 (whitened).]{%
  \includegraphics[width=0.4\textwidth]{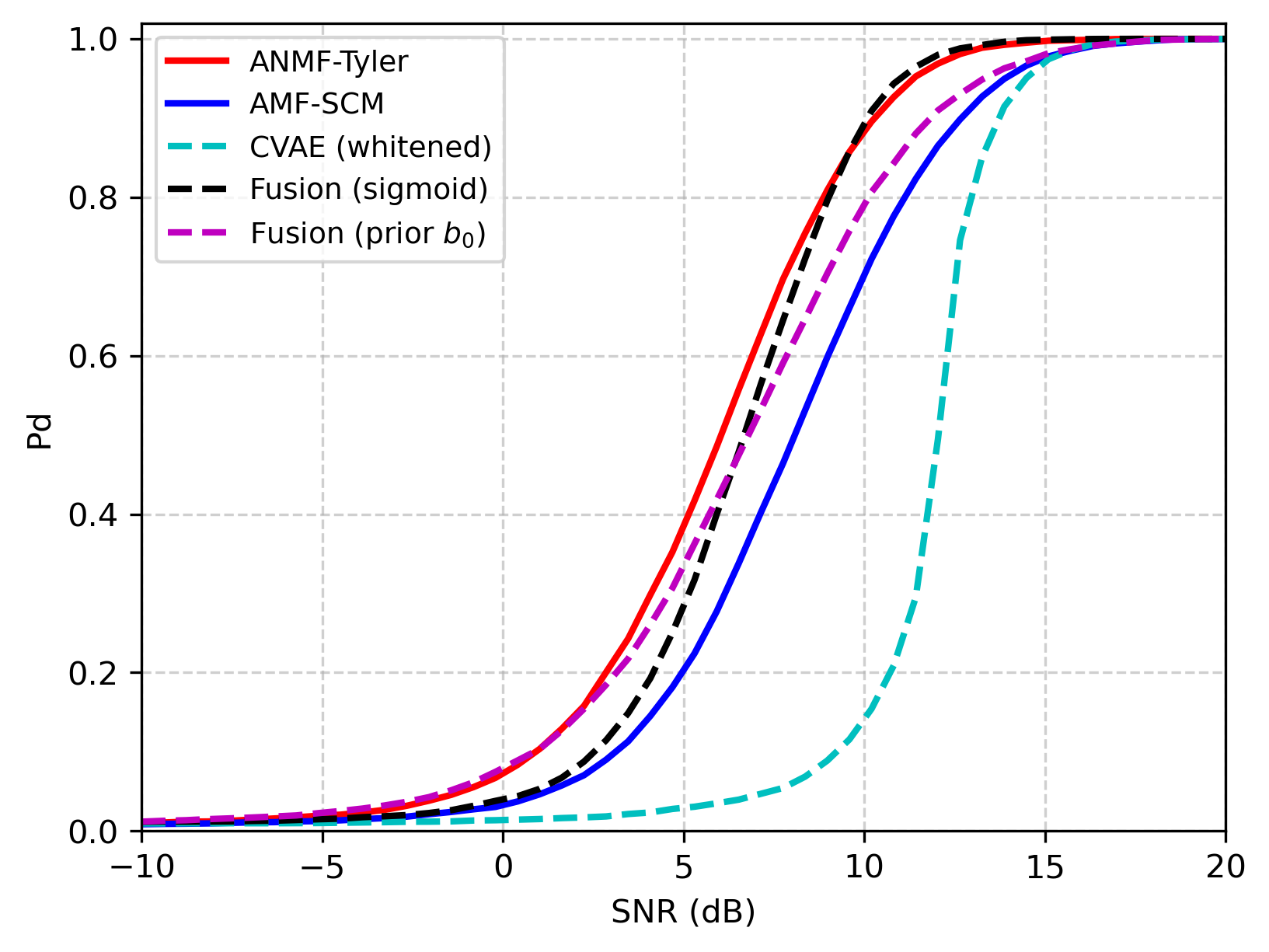}%
  \label{fig:pd_snr_008_white}
}
\caption{$P_d$ vs.\ SNR at $d{=}0$ for CFA16-002 and CFA16-008 ($P_{\mathrm{fa}}=10^{-2}$).
Curves compare AMF-SCM, ANMF-Tyler, CVAE (raw/whitened), and the proposed CVAE+ANMF fusion
(mean-$p$ sigmoid; the $b_0$-prior variant is reported in the appendix).}
\label{fig:pd_snr_d0_all}
\end{figure*}

On CFA16-002 without whitening (Fig.~\ref{fig:pd_snr_d0_all}(a)), ANMF-Tyler is the strongest baseline and outperforms AMF-SCM by about $1$–$1.5$\,dB, while the raw CVAE is clearly miscalibrated and reaches a given $P_d$ only at higher SNR, so that CVAE+ANMF essentially reduces to ANMF. After local whitening (Fig.~\ref{fig:pd_snr_d0_all}(b)), the CVAE becomes more complementary: although not dominant on its own, it improves in the medium-to-high $P_d$ range, and the fusion yields a leftmost curve, typically saving about $1$–$2$\,dB around $P_d\approx 0.8$ compared with ANMF-Tyler.

For CFA16-008, the raw case (Fig.~\ref{fig:pd_snr_d0_all}(c)) shows a similar hierarchy: ANMF-Tyler and AMF-SCM behave as on CFA16-002, whereas the raw CVAE remains weak at low-to-moderate SNR. The sigmoid fusion can then slightly degrade performance with respect to pure ANMF, while the $b_0$-prior fusion tends to track ANMF more closely. Once local whitening is applied (Fig.~\ref{fig:pd_snr_d0_all}(d)), the CVAE strengthens markedly, with a steeper transition in SNR and performance comparable to AMF-SCM; the fusion rules can fully exploit this complementarity, and the sigmoid-based combination becomes the best overall curve, with a gain of roughly $0.5$–$1$\,dB at intermediate $P_d$. Overall, these zero-Doppler results indicate that whitening acts primarily as an enabler for the CVAE, after which CVAE+ANMF can improve upon classical adaptive detectors at fixed CFAR.

\subsection{CSIR Data: Full Doppler $P_d$–SNR Maps}
\label{ssec:alldoppler_csir}

We now examine $P_d$ as a function of SNR and Doppler bin for the real sea clutter. We report separate maps for each scene (CFA16-002 and CFA16-008) and for raw vs.\ locally whitened processing. Each figure stacks AMF-SCM, ANMF-Tyler, CVAE, and the proposed fusion.

\begin{figure*}[!t]
\centering
\includegraphics[width=\textwidth]{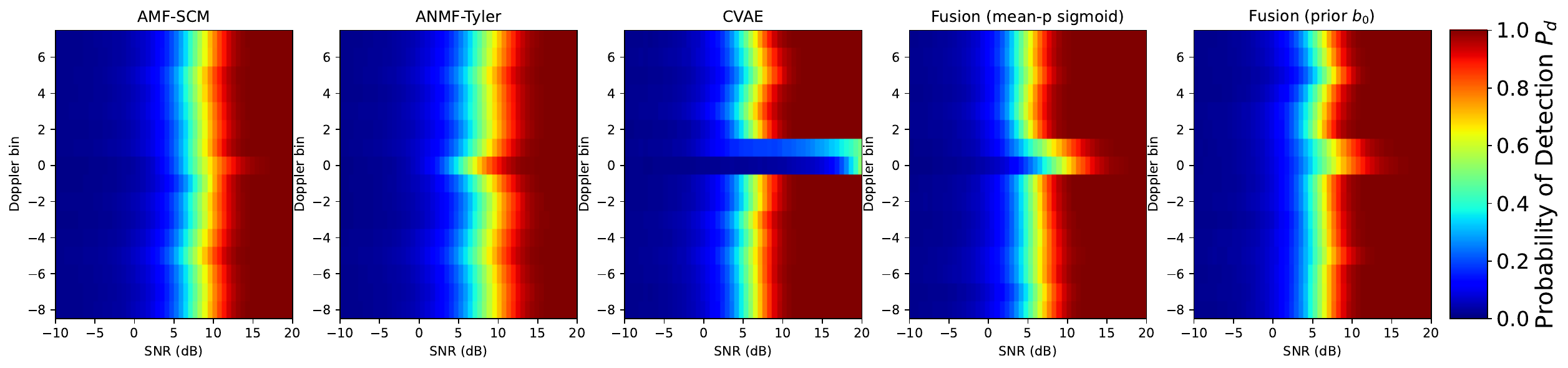}
\caption{$P_d$-SNR-Doppler maps for CFA16-008 (raw, $P_{fa}=10^{-2}$). 
Left to right: AMF-SCM, ANMF-Tyler, CVAE, and proposed fusion (mean-$p$ sigmoid and $b_0$-prior variants).}
\label{fig:maps_008_raw}
\end{figure*}

\begin{figure*}[!t]
\centering
\includegraphics[width=\textwidth]{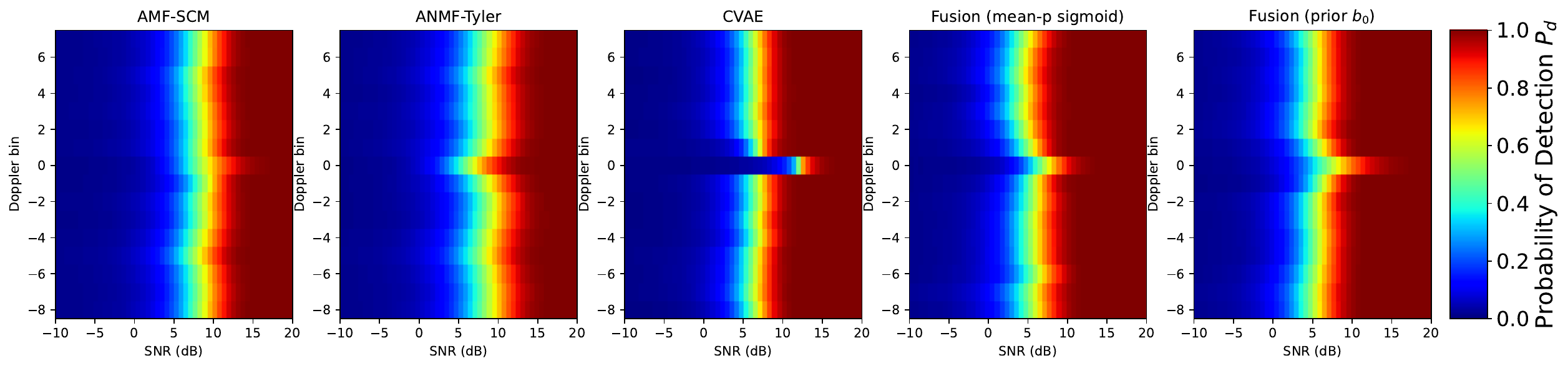}
\caption{$P_d$-SNR-Doppler maps for CFA16-008 (locally whitened).}
\label{fig:maps_008_white}
\end{figure*}

\begin{figure*}[!t]
\centering
\includegraphics[width=\textwidth]{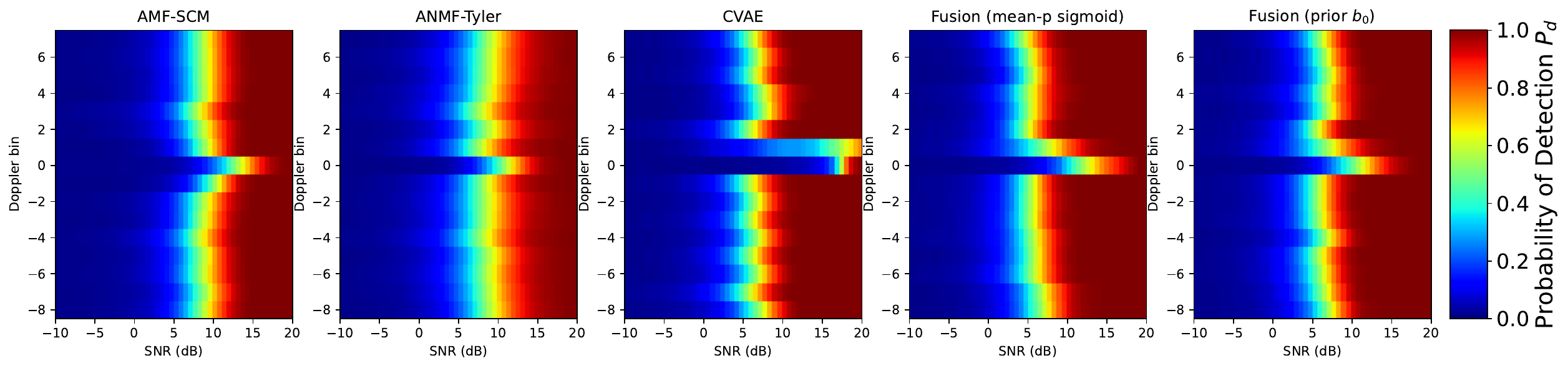}
\caption{$P_d$-SNR-Doppler maps for CFA16-002 (raw).}
\label{fig:maps_002_raw}
\end{figure*}

\begin{figure*}[!t]
\centering
\includegraphics[width=\textwidth]{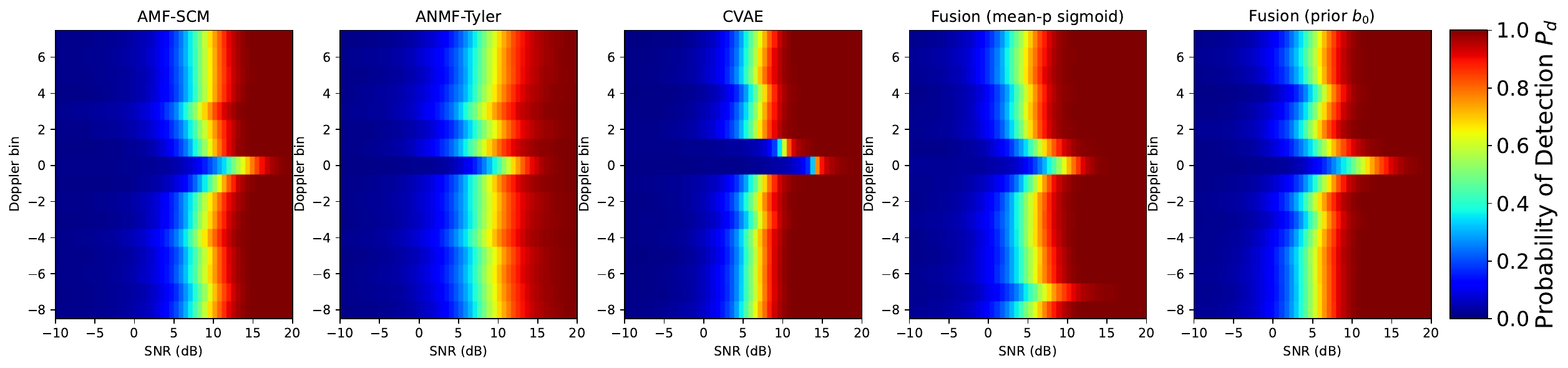}
\caption{$P_d$-SNR-Doppler maps for CFA16-002 (locally whitened).}
\label{fig:maps_002_white}
\end{figure*}

On CFA16-008 with raw clutter (Fig.~\ref{fig:maps_008_raw}), AMF-SCM and ANMF-Tyler yield vertical SNR transition bands, with ANMF slightly better near the clutter ridge ($d\approx 0$). The CVAE achieves a given $P_d$ at lower SNR on most Doppler bins, especially off clutter ridge ($|d|\geq 2$), and the fusion maps exploit this complementarity by tracking CVAE off clutter ridge while staying close to ANMF around $d\in\{-1,0\}$. After whitening (Fig.~\ref{fig:maps_008_white}), the CVAE fronts become more symmetric and shift further left, so that the CVAE dominates the parametric detectors on almost all bins, and the fused detector provides a smooth, high-$P_d$ plateau across Doppler.

For CFA16-002 (Figs.~\ref{fig:maps_002_raw}--\ref{fig:maps_002_white}), the absolute SNR levels change but the qualitative picture is similar: AMF-SCM and ANMF-Tyler show nearly vertical bands with moderate Doppler dependence, the CVAE improves detection on most bins (in particular off clutter ridge), and the fusion acts as an envelope of ANMF and CVAE, closing the small gap at central Doppler while preserving CVAE gains elsewhere. Whitening further homogenizes the maps and smooths the CVAE transitions, notably at low SNR. The qualitative Doppler-central vs.\ off clutter ridge behavior is summarized in Table~\ref{tab:csir_summary}.

\begin{table*}[!t]
\centering
\caption{Qualitative ranking at $d{=}0$ (clutter ridge) on CSIR sea clutter 
($P_{fa}=10^{-2}$). Symbols as in Table~\ref{tab:simu_summary}.
These rankings are restricted to the central Doppler bin; off clutter ridge
($|d|\geq 2$) the CVAE and the proposed fusion generally dominate
ANMF-Tyler, as seen in Figs.~\ref{fig:maps_008_raw}--\ref{fig:maps_002_white}.}
\label{tab:csir_summary}
\begin{tabular}{lccccc}
\toprule
Scene / preprocessing 
& AMF-SCM 
& ANMF-Tyler 
& CVAE 
& Fusion (sigmoid) 
& Fusion ($b_0$-prior) \\
\midrule
CFA16-002, raw      
& +   
& ++  
& --  
& +   
& ++  
\\
CFA16-002, whitened 
& 0   
& +   
& 0   
& ++  
& +   
\\
CFA16-008, raw      
& +   
& ++  
& --  
& 0   
& +   
\\
CFA16-008, whitened 
& +   
& +   
& +   
& ++  
& +   
\\
\bottomrule
\end{tabular}
\end{table*}

\section{Conclusion}
\label{sec:conclu}

We have proposed a complex-valued variational autoencoder for radar detection in non-Gaussian clutter, together with a decision-level fusion rule that combines its output with a model-based ANMF detector. The CVAE is trained only on clutter profiles under $H_0$ and used as an unsupervised generative model; detection relies on a complex reconstruction-error score, calibrated via per-Doppler probability-integral transforms to enforce CFAR. A local whitening scheme was also introduced to stabilize the covariance structure seen by the CVAE and reduce Doppler-wise variability in the null distribution of its scores.

On simulated correlated Gaussian and compound-Gaussian clutter, with and without additional white noise, the CVAE (especially with oracle whitening) closely tracks the MF/NMF baseline in nearly Gaussian regimes and significantly outperforms AMF-SCM and ANMF-Tyler in heavy-tailed or mixed scenarios. These results show that a purely data-driven generative model can recover matched-filter performance when the clutter model is well specified, while offering clear robustness gains when texture and correlation depart from the Gaussian paradigm. On real CSIR sea clutter, the CVAE alone is not uniformly superior to ANMF-Tyler on the clutter ridge, but achieves better detection on most other Doppler bins, particularly after local whitening. The proposed weighted log-$p$ fusion exploits this complementarity: it takes advantage of both detectors namely ANMF-Tyler and CVAE, yielding smoother and more homogeneous $P_d$ surfaces in the Doppler-SNR map while consistently respecting the designed CFAR level.

More broadly, our results illustrate that CVAEs, local covariance normalization, and calibrated $p$-value fusion can be combined to address out-of-distribution detection in non-Gaussian complex-valued signals. 
Promising extensions include moving from one-dimensional Doppler profiles to two-dimensional range-Doppler patches or multi-channel arrays, learning Doppler-dependent fusion weights directly from data while preserving CFAR, and enabling online adaptation of both the CVAE and the null calibration to slowly varying background conditions. Overall, the combination of a complex-valued VAE with classical ANMF through calibrated log-$p$ fusion offers a flexible way to blend model-based and data-driven detection in challenging clutter environments with rigorous false-alarm rate control.

\section*{Acknowledgments}
Part of this work was supported by ANR-ASTRID NEPTUNE 3 (ANR-23-ASM2-0009).
\bibliographystyle{IEEEtran}
\bibliography{refs}

@article{neyman1933problem,
  title={On the Problem of the Most Efficient Tests of Statistical Hypotheses},
  author={Neyman, J. and Pearson, E. S.},
  journal={Philosophical Transactions of the Royal Society of London. Series A},
  volume={231},
  pages={289--337},
  year={1933}
}

@article{ledoit2004well,
  title={A well-conditioned estimator for large-dimensional covariance matrices},
  author={Ledoit, O. and Wolf, M.},
  journal={Journal of Multivariate Analysis},
  volume={88},
  number={2},
  pages={365--411},
  year={2004},
  publisher={Elsevier}
}

@INPROCEEDINGS{7383755,
  author={Ovarlez, J.-P. and Pascal, F. and Breloy, A.},
  booktitle={IEEE 6th International Workshop on Computational Advances in Multi-Sensor Adaptive Processing (CAMSAP)}, 
  title={Asymptotic detection performance analysis of the robust {A}daptive {N}ormalized {M}atched {F}ilter}, 
  year={2015},
  volume={},
  number={},
  pages={137-140},
  doi={10.1109/CAMSAP.2015.7383755}}

@article{Robey1992ACA,
  title={A {CFAR} adaptive matched filter detector},
  author={Robey, F. C. and Fuhrmann, D. R.  and  Kelly, E. J. and  Nitzberg, R.},
  journal={IEEE Transactions on Aerospace and Electronic Systems},
  year={1992},
  volume={28},
  pages={208-216}}

@ARTICLE{6263313,
  author={Ollila, E. and Tyler, D. E. and Koivunen, V. and Poor, H. V.},
  journal={IEEE Transactions on Signal Processing}, 
  title={Complex Elliptically Symmetric Distributions: Survey, New Results and Applications}, 
  year={2012},
  volume={60},
  number={11},
  pages={5597-5625},
  doi={10.1109/TSP.2012.2212433}}

@ARTICLE{4104190,
  author={Kelly, E. J.},
  journal={IEEE Transactions on Aerospace and Electronic Systems}, 
  title={An Adaptive Detection Algorithm}, 
  year={1986},
  volume={AES-22},
  number={2},
  pages={115-127},
  doi={10.1109/TAES.1986.310745}}

@ARTICLE{301849,
  author={Scharf, L. L. and Friedlander, B.},
  journal={IEEE Transactions on Signal Processing}, 
  title={Matched subspace detectors}, 
  year={1994},
  volume={42},
  number={8},
  pages={2146-2157},
  doi={10.1109/78.301849}}

@article{Kraut2001AdaptiveSD,
  title={Adaptive subspace detectors},
  author={Kraut, S. and Scharf, L. L.  and  McWhorter, L. T.},
  journal={IEEE Transactions on Signal Processing},
  year={2001},
  volume={49},
  pages={1-16}
}

@Book{GoodBengCour16,
  Title                    = {Deep Learning},
  Author                   = {Goodfellow, I. J.  and  Bengio, Y. and  Courville, A.},
  Publisher                = {MIT Press},
  Year                     = {2016},
  Address                  = {Cambridge, MA, USA}
}

@INPROCEEDINGS{Kingma_2014,
  author={Kingma, D. P. and Welling, M.},
  booktitle={2nd International Conference on Learning Representations, ICLR}, 
  title={Auto-encoding variational {B}ayes}, 
  year={2014},
  volume={},
  number={},
}

@article{Kingma_2019,
   title={An Introduction to Variational Autoencoders},
   volume={12},
   ISSN={1935-8245},
   DOI={10.1561/2200000056},
   number={4},
   journal={Foundations and Trends® in Machine Learning},
   publisher={Now Publishers},
   author={Kingma, D. P. and Welling, M.},
   year={2019},
   pages={307–392} }

@ARTICLE{MRAMJ2022,
  author={Muzeau, M. and Ren, C. and Angelliaume, S. and Datcu, M. and Ovarlez, J.-P.},
  journal={IEEE Open Journal of Signal Processing}, 
  title={Self-Supervised Learning Based Anomaly Detection in {S}ynthetic {A}perture {R}adar Imaging}, 
  year={2022},
  volume={3},
  number={},
  pages={440-449},
  doi={10.1109/OJSP.2022.3229618}}

@INPROCEEDINGS{marimont2020,
  author={Marimont, S. N. and Tarroni, G.},
  booktitle={IEEE 18th International Symposium on Biomedical Imaging (ISBI)}, 
  title={Anomaly Detection Through Latent Space Restoration Using Vector Quantized Variational Autoencoders}, 
  year={2021},
  volume={},
  number={},
  pages={1764-1767},
  doi={10.1109/ISBI48211.2021.9433778}}

@article{ran2021,
title = {Detecting {O}ut-of-{D}istribution samples via {V}ariational {A}uto-{E}ncoder with reliable uncertainty estimation},
journal = {Neural Networks},
volume = {145},
pages = {199-208},
year = {2022},
issn = {0893-6080},
author = { Ran, X. and  Xu, M. and  Mei, L. and Xu, Q. and Liu, Q.},
}

@article{yang2024,
author={Yang, J.
and Zhou, K.
and Li, Y.
and Liu, Z.},
title={Generalized Out-of-Distribution Detection: A Survey},
journal={International Journal of Computer Vision},
year={2024},
month={Jun},
day={23},
issn={1573-1405},
doi={10.1007/s11263-024-02117-4}
}

@INPROCEEDINGS{KSS2023,
  author={Kahya, S. M. and Yavuz, M. S. and Steinbach, E.},
  booktitle={IEEE International Conference on Acoustics, Speech and Signal Processing (ICASSP)}, 
  title={{MCROOD}: Multi-Class Radar Out-Of-Distribution Detection}, 
  year={2023},
  volume={},
  number={},
  pages={1-5},
  doi={10.1109/ICASSP49357.2023.10095053}}

@INPROCEEDINGS{MSWPNM2021,
  author={Mitiche, I. and Salimy, A. and Werner, F. and Boreham, P. and Nesbitt, A. and Morison, G.},
  booktitle={29th European Signal Processing Conference (EUSIPCO)}, 
  title={{OODCN}: Out-Of-Distribution Detection in Capsule Networks for Fault Identification}, 
  year={2021},
  volume={},
  number={},
  pages={1686-1690},
  doi={10.23919/EUSIPCO54536.2021.9615946}}

@INPROCEEDINGS{BS2023,
  author={Bukhsh, Z. and Saeed, A.},
  booktitle={IEEE International Conference on Acoustics, Speech and Signal Processing (ICASSP)}, 
  title={On Out-of-Distribution Detection for Audio with Deep Nearest Neighbors}, 
  year={2023},
  volume={},
  number={},
  pages={1-5},
  doi={10.1109/ICASSP49357.2023.10094846}}

@article{tyler1987,
	author = "Tyler, D. E.",
	doi = "10.1214/aos/1176350263",
	fjournal = "Annals of Statistics",
	journal = "Ann. Statist.",
	month = "03",
	number = "1",
	pages = "234--251",
	publisher = "The Institute of Mathematical Statistics",
	title = "A Distribution-Free {M}-Estimator of Multivariate Scatter",
	volume = "15",
	year = "1987"
}

@article{pascal08,
	Author = {Pascal, F. and Chitour, Y. and Ovarlez, J.-P. and Forster, P. and Larzabal, P.},
	Journal = {IEEE Transactions on Signal Processing}, 
	Month = {January},
	Pages = {34-48},
	Title = {Covariance Structure Maximum-Likelihood Estimates in Compound {G}aussian Noise: Existence and Algorithm Analysis},
	Volume = {56},
	Year = {2008}}

@ARTICLE{Pascal8,
  author={Pascal, F. and Forster, P. and Ovarlez, J.-P. and Larzabal, P.},
  journal={IEEE Transactions on Signal Processing}, 
  title={Performance Analysis of Covariance Matrix Estimates in Impulsive Noise}, 
  year={2008},
  volume={56},
  number={6},
  pages={2206-2217},
  doi={10.1109/TSP.2007.914311}}

@book{Greco16,
	Editor = {Greco, M. S.  and De Maio, A.},
	Month = {Jan},
	Publisher = {SciTech Publishing},
	Title = {Modern Radar Detection Theory},
	Year = {2016}}

@inproceedings{nakashika20_interspeech,
  title     = {Complex-Valued Variational Autoencoder: A Novel Deep Generative Model for Direct Representation of Complex Spectra},
  author    = {Toru Nakashika},
  year      = {2020},
  booktitle = {Interspeech 2020},
  pages     = {2002--2006},
  doi       = {10.21437/Interspeech.2020-1964},
  issn      = {2958-1796},
}

@INPROCEEDINGS{deepcomplexconvxiangyang,
  author={Xiang, Y. and Tian, J. and Hu, X. and Xu, X. and Yin, Z.},
  booktitle={IEEE International Conference on Acoustics, Speech and Signal Processing (ICASSP)}, 
  title={A Deep Representation Learning-Based Speech Enhancement Method Using Complex Convolution Recurrent Variational Autoencoder}, 
  year={2024},
  volume={},
  number={},
  pages={781-785},
  doi={10.1109/ICASSP48485.2024.10448125}}

@INPROCEEDINGS{complexreccurrxie,
  author={Xie, Y. and Arildsen, T. and Tan, Z.-H.},
  booktitle={International Joint Conference on Neural Networks (IJCNN)}, 
  title={Complex Recurrent Variational Autoencoder for Speech Resynthesis and Enhancement}, 
  year={2024},
  volume={},
  number={},
  pages={1-7},
  doi={10.1109/IJCNN60899.2024.10650194}}

@article{barrachina:hal-03771786,
  TITLE = {{Comparison Between Equivalent Architectures of Complex-valued and Real-valued Neural Networks - Application on Polarimetric {SAR} Image Segmentation}},
  AUTHOR = {Barrachina, J. A. and Ren, C. and Morisseau, C. and Vieillard, G. and Ovarlez, J.-P.},
  JOURNAL = {{Journal of Signal Processing Systems}},
  PUBLISHER = {{Springer}},
  YEAR = {2022},
  MONTH = Jul,
  DOI = {10.1007/s11265-022-01793-0}
}

@INPROCEEDINGS{papierICASSP,
  author={Rouzoumka, Y. A. and Terreaux, E. and Morisseau, C. and Ovarlez, J.-P. and Ren, C.},
  booktitle={IEEE International Conference on Acoustics, Speech and Signal Processing (ICASSP)}, 
  title={Out-of-Distribution Radar Detection in Compound
Clutter and Thermal Noise through Variational
Autoencoders}, 
  year={2025},
  volume={},
  number={},
  pages={},
  doi={}
}

@InProceedings{pmlr-v172-graham22a,  
title =  {Transformer-based out-of-distribution detection for clinically safe segmentation},  
author =       {Graham, M. S and Tudosiu, P.-D. and Wright, P. and Pinaya, W. H. L. and U-King-Im, J.-M. and Mah, Y. H and Teo, J. T and Jager, R. and Werring, D. and Nachev, P. and Ourselin, S. and Cardoso, M. J.},  
booktitle =  {Proceedings of The 5th International Conference on Medical Imaging with Deep Learning},  
pages =  {457--476},  
year =  {2022},  
volume =  {172},  
series =  {Proceedings of Machine Learning Research},  month =  {06--08 Jul},  
publisher =    {PMLR}}

@article{Xu2021AnomalyTT,
  title={Anomaly Transformer: Time Series Anomaly Detection with Association Discrepancy},
  author={ Xu, J. and  Wu, H. and  Wang, J. and  Long, M.},
  journal={ArXiv},
  year={2021},
  volume={abs/2110.02642}
}

@inproceedings{NEURIPS2020_41d80bfc,
 author = {Wu, H. and K\"{o}hler, J. and Noe, F.},
 booktitle = {Advances in Neural Information Processing Systems},
 editor = {H. Larochelle and M. Ranzato and R. Hadsell and M. F. Balcan and H. Lin},
 pages = {5933--5944},
 publisher = {Curran Associates, Inc.},
 title = {Stochastic Normalizing Flows},
 volume = {33},
 year = {2020}
}

@inproceedings{NEURIPS2020_9578a63f,
 author = {Nielsen, D. and Jaini, P. and Hoogeboom, E. and Winther, O. and Welling, M.},
 booktitle = {Advances in Neural Information Processing Systems},
 editor = {H. Larochelle and M. Ranzato and R. Hadsell and M.F. Balcan and H. Lin},
 pages = {12685--12696},
 publisher = {Curran Associates, Inc.},
 title = {{SurVAE} Flows: Surjections to Bridge the Gap between {VAE}s and Flows},
 volume = {33},
 year = {2020}
}

@article{tuli2022tranad,
  title={{TranAD: Deep Transformer Networks for Anomaly Detection in Multivariate Time Series Data}},
  author={Tuli, S. and Casale, G. and Jennings, N. R.},
  journal={Proceedings of VLDB},
  volume={15},
  number={6},
  pages={1201-1214},
  year={2022}
}

@inproceedings{NEURIPS2020_f5496252,
 author = {Liu, W. and Wang, X. and Owens, J. and Li, Y.},
 booktitle = {Advances in Neural Information Processing Systems},
 editor = {H. Larochelle and M. Ranzato and R. Hadsell and M. F. Balcan and H. Lin},
 pages = {21464--21475},
 publisher = {Curran Associates, Inc.},
 title = {Energy-based Out-of-distribution Detection},
 volume = {33},
 year = {2020}
}

@INPROCEEDINGS{9761434,
  author={Li, X. and Desrosiers, C. and Liu, X.},
  booktitle={2022 IEEE 19th International Symposium on Biomedical Imaging (ISBI)}, 
  title={Symmetric Contrastive Loss for Out-of-Distribution Skin Lesion Detection}, 
  year={2022},
  volume={},
  number={},
  pages={1-5},
  doi={10.1109/ISBI52829.2022.9761434}}

@inproceedings{zhou-etal-2021-contrastive,
    title = "Contrastive Out-of-Distribution Detection for Pretrained Transformers",
    author = "Zhou, W.  and
      Liu, F.  and
      Chen, M.",
    booktitle = "Proceedings of the 2021 Conference on Empirical Methods in Natural Language Processing",
    month = nov,
    year = "2021",
    publisher = "Association for Computational Linguistics",
    doi = "10.18653/v1/2021.emnlp-main.84",
    pages = "1100--1111"
}

@article{Baur_2021,
title = {Autoencoders for unsupervised anomaly segmentation in brain {MR} images: A comparative study},
journal = {Medical Image Analysis},
volume = {69},
pages = {101952},
year = {2021},
issn = {1361-8415},
doi = {https://doi.org/10.1016/j.media.2020.101952},
author = { Baur, C. and  Denner, S. and  Wiestler, B. and Nassir Navab and  Albarqouni, S.}
}

@article{Rosenblatt1952,
author = {Rosenblatt, M.},
title = {{Remarks on a Multivariate Transformation}},
volume = {23},
journal = {The Annals of Mathematical Statistics},
number = {3},
publisher = {Institute of Mathematical Statistics},
pages = {470 -- 472},
year = {1952},
doi = {10.1214/aoms/1177729394}
}

@book{Fisher1932,
  author    = {Fisher, R. A.},
  title     = {Statistical Methods for Research Workers},
  edition   = {4th},
  publisher = {Oliver and Boyd},
  address   = {Edinburgh},
  year      = {1932}
}

@article{Lancaster1961,
author = {Lancaster, H. O.},
title = {THE COMBINATION OF PROBABILITIES: AN APPLICATION OF ORTHONORMAL FUNCTIONS},
journal = {Australian Journal of Statistics},
volume = {3},
number = {1},
pages = {20-33},
doi = {https://doi.org/10.1111/j.1467-842X.1961.tb00058.x},
year = {1961}
}

@article{Diebold1998,
 ISSN = {00206598, 14682354},
 author = {Diebold, F. X.  and Gunther, T. A.  and  Tay, A. S.},
 journal = {International Economic Review},
 number = {4},
 pages = {863--883},
 publisher = {[Economics Department of the University of Pennsylvania, Wiley, Institute of Social and Economic Research, Osaka University]},
 title = {Evaluating Density Forecasts with Applications to Financial Risk Management},
 urldate = {2025-10-03},
 volume = {39},
 year = {1998}
}

@article{Gneiting2007,
author = {Gneiting, T. and Raftery, A. E.},
address = {Alexandria, VA},
copyright = {American Statistical Association 2007},
issn = {0162-1459},
journal = {Journal of the American Statistical Association},
language = {eng},
number = {477},
pages = {359-378},
publisher = {Taylor & Francis},
title = {Strictly Proper Scoring Rules, Prediction, and Estimation},
volume = {102},
year = {2007},
}

@inproceedings{fixtorchcvnn,
  TITLE = {{$\mathrm{Torchcvnn}$: A PyTorch-based library to easily experiment with state-of-the-art Complex-Valued Neural Networks}},
  AUTHOR = {Fix, J. and Gabot, Q. and Huy N., X and Frontera-Pons, J. and Ren, C. and Ovarlez, J.-P.},
  BOOKTITLE = {{International Joint Conference on Neural Networks}},
  ADDRESS = {Rome, Italy},
  YEAR = {2025},
  MONTH = Jun
}

@inproceedings{rouzoumkacvae2025,
  TITLE = {{Complex-Valued Variational Autoencoders for Radar Detection in Joint Compound Gaussian Clutter and Thermal Noise}},
  AUTHOR = {Rouzoumka, Y. A. and Terreaux, E. and Morisseau, C. and Ovarlez, J.-P and Ren, C.},
  BOOKTITLE = {{EUSIPCO 2025 -European Signal Processing Conference}},
  ADDRESS = {Palerme, Italy},
  YEAR = {2025},
  MONTH = Sep
}

@article{deWind2010DataWare,
  author    = {H.~J.~de Wind and J.~E.~Cilliers and P.~L.~Herselman},
  title     = {DataWare: Sea Clutter and Small Boat Radar Reflectivity Databases},
  journal   = {IEEE Signal Processing Magazine},
  year      = {2010},
  volume    = {27},
  number    = {2},
  pages     = {145--148},
  doi       = {10.1109/MSP.2009.935415}
}

@inproceedings{Herselman2007Analysis,
  author    = {P.~L.~Herselman and C.~J.~Baker},
  title     = {Analysis of calibrated sea clutter and boat reflectivity data at C- and X-band in South African coastal waters},
  booktitle = {IET International Conference on Radar Systems, 2007},
  year      = {2007},
  doi       = {10.1049/cp:20070616}
}

@article{Tan2025GNNSeaClutter,
  author    = {J.~Tan and W.~Sheng and H.~Zhu},
  title     = {A Sea Clutter Suppression Method with Graph Neural Network for Maritime Target Detection},
  journal   = {IEEE Transactions on Aerospace and Electronic Systems},
  year      = {2025},
  note      = {Early Access}}

@article{Liu2021WaveHeight,
  author    = {N.~Liu and X.~Jiang and H.~Ding and Y.~Xu},
  title     = {Wave Height Inversion and Sea State Classification Based on Deep Learning of Radar Sea Clutter Data},
  journal   = {IEEE Geoscience and Remote Sensing Letters},
  year      = {2021},
  doi       = {10.1109/LGRS.2021.3114549}
}

@inproceedings{Gong2020LightCNN,
  author    = {J.~Gong and A.~Wang and W.~Chen},
  title     = {LightCNN: A Compact CNN for Moving Maritime Targets Detection},
  booktitle = {IEEE Radar Conference},
  year      = {2020},
  doi       = {10.1109/RadarConf2043947.2020.9320939}
}

@inproceedings{Ding2020MarineDataset,
  author    = {H.~Ding and N.~Liu and W.~Zhou and Y.~Xue and J.~Guan},
  title     = {Construction of Marine Target Detection Dataset for Intelligent Radar Application},
  booktitle = {International Conference on Intelligent Computing},
  year      = {2020},
  pages     = {333--344},
  doi       = {10.1007/978-981-15-0187-6_28}
}

@article{Ma2019SeaClutterDL,
  author    = {L.~Ma and J.~Wu and J.~Zhang and Z.~Wu and G.~Jeon},
  title     = {Research on Sea Clutter Reflectivity Using Deep Learning Model in Industry 4.0},
  journal   = {IEEE Access},
  year      = {2019},
  volume    = {7},
  pages     = {171652--171662},
  doi       = {10.1109/ACCESS.2019.2955963}
}

@article{Kandagatla2025NNSeaClutter,
  author    = {R.~K.~Kandagatla and V.~Malapati and K.~Cherukuri},
  title     = {Sea Clutter Suppression Using Neural Network},
  journal   = {Marine Technology Research},
  year      = {2025},
  doi       = {10.33175/mtr.2025.272334}}

@ARTICLE{HOOD,
  author={Kahya, Sabri Mustafa and Yavuz, Muhammet Sami and Steinbach, Eckehard},
  journal={IEEE Transactions on Radar Systems}, 
  title={{HOOD}: Real-Time Human Presence and Out-of-Distribution Detection Using {FMCW} Radar}, 
  year={2025},
  volume={3},
  number={},
  pages={44-56},
  doi={10.1109/TRS.2024.3514840}}

@article{dkw_massart1990,
 ISSN = {00911798, 2168894X},
 author = {P. Massart},
 journal = {The Annals of Probability},
 number = {3},
 pages = {1269--1283},
 publisher = {Institute of Mathematical Statistics},
 title = {The Tight Constant in the {D}voretzky-{K}iefer-{W}olfowitz Inequality},
 urldate = {2025-11-19},
 volume = {18},
 year = {1990}
}

@ARTICLE{picinbono1996complexdistribution,
  author={Picinbono, B.},
  journal={IEEE Transactions on Signal Processing}, 
  title={Second-order complex random vectors and {N}ormal distributions}, 
  year={1996},
  volume={44},
  number={10},
  pages={2637-2640},
  doi={10.1109/78.539051}}

@book{SchreierScharf2010,
    place={Cambridge},
    title={Statistical Signal Processing of Complex-Valued Data: The Theory of Improper and Noncircular Signals},
    publisher={Cambridge University Press},
    author={Schreier, Peter J. and Scharf, Louis L.},
    year={2010}}


 





\end{document}